\documentclass[lettersize,journal]{IEEEtran}
\usepackage{amsmath,amsfonts}
\usepackage{algorithmic}
\usepackage{algorithm}
\usepackage{array}
\usepackage[caption=false,font=normalsize,labelfont=sf,textfont=sf]{subfig}
\usepackage{textcomp}
\usepackage{stfloats}
\usepackage{url}
\usepackage{verbatim}
\usepackage{graphicx}
\usepackage{cite}
\usepackage{enumitem}
\usepackage{pifont}
\usepackage{hyperref}
\begin{document}
\title{Data Optimization in Deep Learning: A Survey}

\author{Ou Wu, Rujing Yao
\thanks{Ou Wu is with National Center for Applied Mathematics, Tianjin University, Tianjin,
China, 300072. Rujing Yao is with Department of Information Resources Management, Nankai University, Tianjin,
China, 300071. E-mail: wuou@tju.edu.cn, rjyao@mail.nankai.edu.cn.}
}

\maketitle

\begin{abstract}
Large-scale, high-quality data are considered an essential factor for the successful application of many deep learning techniques. Meanwhile, numerous real-world deep learning tasks still have to contend with the lack of sufficient amounts of high-quality data. Additionally, issues such as model robustness, fairness, and trustworthiness are also closely related to training data. Consequently, a huge number of studies in the existing literature have focused on the data aspect in deep learning tasks. Some typical data optimization techniques include data augmentation, logit perturbation, sample weighting, and data condensation. These techniques usually come from different deep learning divisions and their theoretical inspirations or heuristic motivations may seem unrelated to each other. This study aims to organize a wide range of existing data optimization methodologies for deep learning from the previous literature, and makes the effort to construct a comprehensive taxonomy for them. The constructed taxonomy considers the diversity of split dimensions, and deep sub-taxonomies are constructed for each dimension. On the basis of the taxonomy, connections among the extensive data optimization methods for deep learning are built in terms of four aspects. We probe into rendering several promising and interesting future directions. The constructed taxonomy and the revealed connections will enlighten the better understanding of existing methods and the design of novel data optimization techniques. Furthermore, our aspiration for this survey is to promote data optimization as an independent subdivision of deep learning. A curated, up-to-date list of resources related to data optimization in deep learning is available at \url{https://github.com/YaoRujing/Data-Optimization}.
\end{abstract}


\section{Introduction}\label{section1}
\IEEEPARstart{D}{eep} learning has received increasing attention in both the AI community and many application domains due to its superior performance in various machine-learning tasks in recent years. A successful application of deep learning cannot leave the main factors, which include a properly designed deep neural network~(DNN), a set of high-quality training data, and a well-suited learning strategy~(e.g., initialization schemes for hyper-parameters). Among the main factors, training data is of great importance and it usually plays a decisive role in the entire training process~\cite{DCA2021}. The concept of data-centric AI is rising, which breaks away from the widespread model-centric perspective~\cite{TPO2023}. Large models like GPT-4 show significant potential in the direction of achieving general artificial intelligence~(AGI). It is widely accepted that the training for large models requires a huge size of high-quality training data.

However, most real applications lack ideal training data. Real training data usually encounters one or several of the nine common issues as shown in Fig.~\ref{fig1}. The following six issues are directly related to training data:
\begin{itemize}
    \item[(1)] Biased distribution: This issue denotes that the distribution of the training data does not conform to the true distribution of the data in a learning task. One typical bias is class imbalance, in which the proportions of different categories in the training data are not identical due to reasons such as data collection difficulties, whereas the proportions of different categories in test data are identical. 
    \item[(2)] Low quality: This issue corresponds to at least two scenarios. The first refers to data noise that either partial training samples or partial training labels contain noises. As for sample noises, partial samples themselves are corrupted by noises. Taking optical character recognition~(OCR) for example, some scanned images may contain serious background noises. The second typical case occurs in multi-model/multi-view learning scenarios. Inconsistency and information missing may exist~\cite{MGL2022, LHG2023}. For instance, the text title for an image may be mistakenly provided, or it may contain limited words with little information.  \begin{figure}[t] 
    \centering \includegraphics[width=0.8\linewidth]{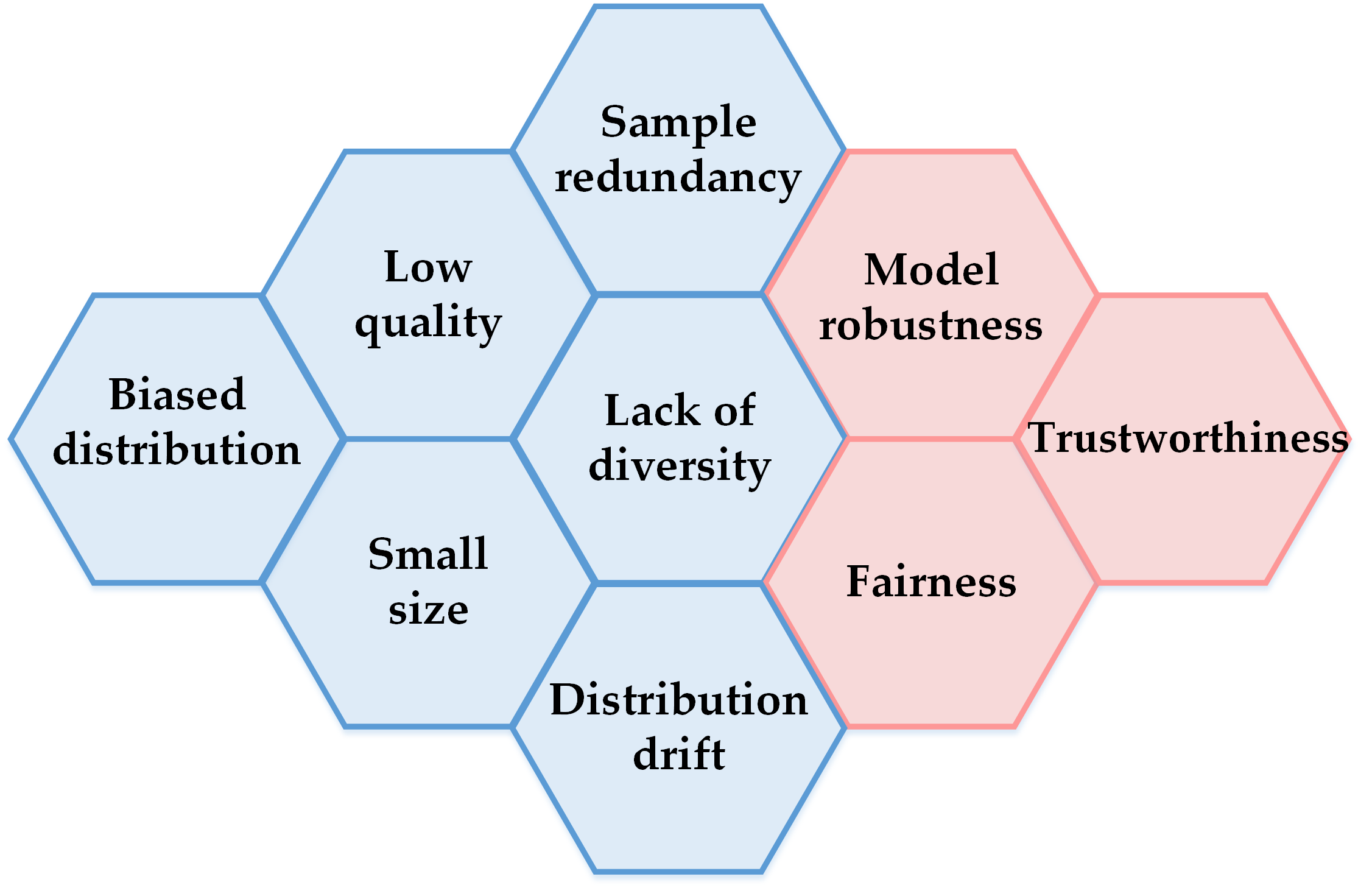}    \vspace{-0.12in}
    \caption{Nine issues around real training data.}
    \label{fig1}
     \vspace{-0.1in}
\end{figure}
    \item[(3)] Small size: The training size surely impacts the training performance~\cite{ACL2021}. The larger the training data, the better the training performance usually being attained. Due to insufficient data collection budget or technique limitation, the training data will be relatively small for real use. Therefore, learning under small-size training data is a serious concern in deep learning. This study does not discuss the extreme cases of small size, such as few/one/zero-shot learning.

    \item[(4)] Sample redundancy: Even though large training data is expected, it does not mean that every datum is useful. There are still learning tasks that the training set contains redundant data~\cite{SRI2019}. Two typical cases exist. First, the training size is relatively large and exceeds the processing capacity of the available computing hardware. Second, some regions of training samples may be sampled excessively, and the deletion of such excessive training samples does not affect the training performance. In this case, sample redundancy may occur in certain subsets of some categories.

    \item[(5)] Lack of diversity: This issue refers to the fact that some attributes for certain categories concentrate excessively in the training corpus. Data diversity is also crucial for DNN training~\cite{CDD2022}. Taking object classification as an example, the backgrounds in images of the dog category may usually be green grass. However, the ``dog" category is not necessarily related to green grass. The lack of diversity in some non-essential attributes can lead to a spurious correlation between some non-essential attributes and the category. This issue is similar to the second case of sample redundancy. Nevertheless, lack of diversity does not necessarily imply the presence of redundant samples.
    \item[(6)] Distribution drift: This issue denotes that the distribution of the involved data varies over time. Indeed, distribution drift may occur in most real learning applications, as either the concept or the form (e.g., object appearances, text styles) of samples varies fast or slow. Concept drift~\cite{LUC2018} is the research focus in distribution drift. 
\end{itemize}
The above summary of data issues is not mutually exclusive, as there are overlaps among different issues. For example, small size may only occur in several categories, which can also be attributed to a type of biased distribution. Besides these data issues, there are also some other~(not exhaustive) issues closely related to the training data: 

\begin{itemize}
    \item[(7)] Model robustness: This issue concerns the resistance ability of a DNN model to adversarial attacks~\cite{TARD2023}. Model robustness is highly important for applications related to health, finance, and human life. If DNN models for these applications are compromised by adversarial attacks, serious consequences may ensue.  

    \item[(8)] Fairness: This issue concerns the performance differences among different categories or different attributes in a learning task~\cite{FCC2018}. For example, the recognition accuracy of faces in different color groups should be at the same level.
    
    \item[(9)] Trustworthiness. This issue has emerged in recent years as deep learning has been gradually applied in many safety-critical applications such as autonomous driving and medical assistance~\cite{TAR2022}. This issue is closely related to robustness and fairness. It mainly refers to the explainability and calibration of DNN models.  
\end{itemize}

To address the above-mentioned issues, numerous theoretical explorations have been conducted and tremendous new methodologies have been proposed in previous literature. Most of these existing methods directly optimize the involved data in learning rather than explore new DNN structures, which is referred to as \textbf{data optimization} for deep learning in this paper. As the listed issues belong to different machine learning divisions, the inspirations and focuses of these methods are usually distinct and seem unrelated to each other. For instance, the primary learning strategy for imbalanced learning (belonging to the biased distribution issue) is sample weighting which assigns different weights to training samples in deep learning training epochs. The primary manipulation for the small-size issue is to employ the data augmentation technique such as image resize and mixup~\cite{MBE2018} for image classification. When dealing with label noise in deep learning, one strategy is to identify noisy labels and then remove them during training. In cases where training data for certain categories lack sufficient diversity, causal learning is employed to break down the spurious correlations among labels and some irrelevant attributes such as certain backgrounds. Due to the apparent lack of connection, these studies typically do not mutually cite or discuss each other.

Our previous study~\cite{CL2022} partially reveals that one technique, namely, data perturbation, has been leveraged to deal with most aforementioned issues. This observation illuminates us to explore the data optimization methodologies for those issues in a more broad view. In this study, a comprehensive review for a wide range of data optimization methods is conducted. First, a systematic data optimization taxonomy is established in terms of eight dimensions, including pipeline, object, technical path, and so on. Second, the intrinsic connections among some classical methods are explored according to four aspects, including data perception, application scenario, similarity/opposite, and theory. Third, theoretical studies are summarized for the existing data optimization techniques. Lastly, several future directions are presented according to our analysis. 

The differences between our survey and existing surveys in relevant areas, including imbalanced learning, noisy-label learning, data augmentation, adversarial training, and distillation, lie in two aspects. First, this survey takes a data-centric view for studies from a wide range of distinct deep learning realms. Therefore, our focus is merely on the data optimization studies for the listed issues. Methods that do not belong to data optimization for the listed issues are not referred to in this study. Second, the split dimensions~(e.g., data perception and theory) which facilitate the establishment of connections among seemingly unrelated methods are considered in our taxonomy. These dimensions are usually not referred to in the existing surveys.

The contributions of this study are summarized as follows. 
\begin{itemize}
    \item Methodologies related to data enhancement for dealing with distinct deep learning issues are reviewed with a new taxonomy. To our knowledge, this is the first work that aims to construct a data-centric taxonomy focusing on data optimization across multiple deep learning divisions and applications. 
    \item The connections among many seemingly unrelated methods are built according to our constructed taxonomy. The connections can inspire researchers to design more potential new techniques. 
    \item Theoretical studies for data optimization are summarized and interesting future directions are discussed.
\end{itemize} 

This paper is organized as follows. Section II introduces main survey studies related to data optimization techniques. Section III describes the main framework of our constructed data optimization taxonomy. Sections IV, V, VI, and VII introduce the details of our taxonomy. Section VIII explores the connections among different data optimization techniques. Section IX presents several future directions, and conclusions are presented in Section X.
\section{Related studies}
The issues listed in the previous section gradually spawn numerous independent research realms of deep learning. Subsequently, there have been many survey studies conducted for these issues. The following introduces related surveys in several typical research topics.

\textbf{Imbalanced learning.} It is a hot research area in deep learning~\cite{DGM2023}. He and Garcia~\cite{he2009learning} conducted the first comprehensive yet deep survey study on imbalanced learning. They explored the intrinsic characteristics of learning tasks incurred by imbalanced data. It is noteworthy that He and Garcia pointed out that an imbalanced dataset is ``a high-complexity dataset with both between-class and within-class imbalances, multiple concepts, overlapping, noise, and a lack of representative data". This statement refers to most data issues listed in Section I. For instance, the lack of diversity and sample redundancy can be considered as a lack of representativeness. Recent studies have focused on the extreme case of imbalanced learning, namely, long-tailed classification. Zhang et al.~\cite{zhang2023deep} summarized the recent developments in deep long-tailed classification. In their constructed taxonomy, module improvement such as a new classifier is listed as one of the three main techniques. In this study, module improvement is not considered, as it does not fall under data optimization.

\textbf{Noisy-label learning.} This is another research area gaining tremendous attention in recent years as label noise is nearly unavoidable in real learning tasks. Algan and Ulisory~\cite{ICW2021} summarized the methods in noisy-label learning for image classification. Song et al.~\cite{LFN2022} elaborately designed taxonomy for noisy-label learning along with three categories, including ``data", ``objective", and ``optimization". Their taxonomy facilitates the understanding of a huge number of existing techniques. Nevertheless, overlap exists between the three dimensions. For example, reweighting locates in the ``objective" category, whereas learning to reweight locates in the ``optimization" category. The taxonomy introduces in this study may aid the construction a more appropriate taxonomy for noisy-label learning.

\textbf{Learning with small data}. Big data has achieved great success in deep learning tasks. Meanwhile, many real learning tasks still confront with the challenge of small-size training data. Cao et al.~\cite{ASO2022} performed rigorous theoretical analysis for the generalization error and label complexity of learning on small data. They categorized the small-data learning methods into those with the Euclidean or non-Euclidean mean representation. Wang et al.~\cite{GFA2020} constructed a few-shot learning taxonomy with three folds, including ``data", ``model", and ``algorithm". Data-centric learning methods are also among the primary choices for few-shot learning. 

\textbf{Concept drift}. Lu et al.~\cite{LUC2018} investigated the learning for concept drift under three components, including concept drift detection, concept drift understanding, and concept drift adaptation. Yuan et al.~\cite{RAI2022} divided existing studies into two categories, namely, model parameter updating and model structure updating in concept drift adaptation. This division is from the viewpoint of the model. Indeed, pure data-based strategy is also employed in learning under concept drift. For example, Diez-Olivan et al.~\cite{ADC2021} leveraged data augmentation to fine-tune the last layer of DNNs for quickly concept drift adaptation. This study may motivate researchers on distribution drift to focus more on data optimization manners. Some early surveys can be found in~\cite{ASOC2014, AOOC2019}.
 
\textbf{Adversarial robustness}. In many studies, model robustness is limited to adversarial robustness. Silva and Najafirad~\cite{OAC2020} explored challenges and future directions for model robustness in deep learning. They divided existing adversarial robust learning methods into three categories, including adversarial training, regularization, and certified defenses. Xu et al.~\cite{ROD2021} summarized the studies for model robustness on graphs. Goyal et al.~\cite{ASI2023} reviewed the adversarial defense and robustness in the NLP community. Their constructed taxonomy contains four categories, including adversarial training, perturbation control, certification, and miscellaneous. In this study, adversarial training is taken as a data optimization strategy, as it enhances the training set by adding or virtually adding new data.

\textbf{Fairness-aware learning}. It receives increasingly attention in recent years. Mehrabi et al.~\cite{ASOB2021} explored different sources of biases that can affect the fairness of learning models. They revealed that each of the three factors, namely, data, learning algorithms, and involved users may result in bias. Petrović et al.~\cite{FFA2022} pointed out that sample reweighting and adversarial training are two common strategies for fair machine learning. 

\textbf{Trustworthy learning}. It is the key of trustworthy AI, which aims to ensure that an AI system is worthy of being trusted. Trust is a complex phenomenon~\cite{TOA2018} highly related to fairness, explainability, reliability, etc. Kaur et al.~\cite{TAI2022} summarized studies on trustworthy artificial intelligence in a quite broad view. Wu et al.~\cite{TGL2022} provided an in-depth review for studies about trustworthy learning on graphs.

There are also studies that focus on learning tasks with more than one of the listed data issues. For example, Fang et al.~\cite{CNL2022} addressed noisy-label learning under the long-tailed distributions of training data. Singh et al.~\cite{AES2021} conducted an empirical study concerning fairness, adversarial robustness, and concept drift, simultaneously. To our knowledge, no survey study pays attention to the intersection of the research areas related to the listed issues. The unified taxonomy constructed in this survey would enlighten the study on the intersection of multiple research areas.

The most similar study to this work is the survey presented by Wan et al.~\cite{ASODO2022}, which focuses on data optimization in computer vision. There are significant differences between our and Wan et al.'s study. First, the covered technical scopes of ours are much broader than those of Wan et al.'s study. Their study limits the scope merely in data selection, including resampling, subset selection, and active learning-based selection. Nevertheless, perturbation, weighting, dataset distillation, and augmentation which attempt to optimize the training data without modifying the backbone network are considered as the data optimization in this study. Second, the split dimensions of ours are quite different from those in Wan et al.'s study for the overlapped methods. For example, curriculum learning is divided into the resampling category, whereas it is divided into the weighting category in our study. Dataset distillation is merged into one division, namely, data pruning. In contrast, dataset distillation is not included in Wan et al.'s study. Lastly, additional important parts including data perception, connections among different paths, and theoretical investigation are introduced and discussed in this study. Zheng et al.~\cite{TDC2023} holds a data-centric perspective to review studies on graph machine learning, which is also similar in spirit with our study. Most data issues summarized in this study are also discussed in their study. They divided existing studies into graph data collection, enhancement, exploration, maintenance~(for privacy and security), and graph operations, which are not applicable for our taxonomy.

It is noteworthy that classical data pre-processing methodologies such as data cleaning~(e.g., missing data imputation), standardization~(e.g., z-score), and transformation~(e.g., data discretization) also aim to make data better for learning. Considering that these methodologies are mature and mainly utilized in shallow learning, they are not introduced in this survey.

\begin{figure*}[t] 
    \centering \includegraphics[width=0.8\linewidth]{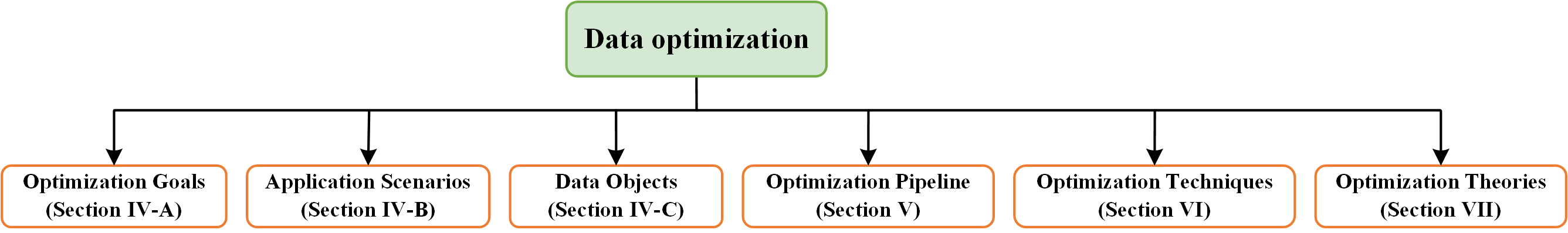}    \vspace{-0.12in}
    \caption{The six split dimensions of our constructed taxonomy for data optimization.}
    \label{fig2}
     \vspace{-0.16in}
\end{figure*}

\section{Overall of The proposed taxonomy}
To ensure our constructed taxonomy well organized and comprehensive coverage on previous data optimization techniques about the issues listed in Section I as much as possible, the following principles for the design of split dimensions are considered.
\begin{itemize}
    \item[(1)] The first layer of the taxonomy should consider multiple views, with each view corresponding to a sub-taxonomy. Most existing taxonomies for specific research realms adopt only a single view. In this study, only a single view is inadequate for systematically arranging studies from various deep learning realms.
    \item[(2)] The dividing dimension should be general so as to embrace existing studies as much as possible. Therefore, the dimensions designed in existing taxonomies for specific research areas should not be directly followed. A new comprehensive taxonomy is required.
    \item[(3)] The new taxonomy should be compatible with existing taxonomies. That is, inconsistency between our and existing taxonomies is allowed. However, contradiction between them should be avoided.
\end{itemize}

\begin{figure}[t] 
    \centering \includegraphics[width=1\linewidth]{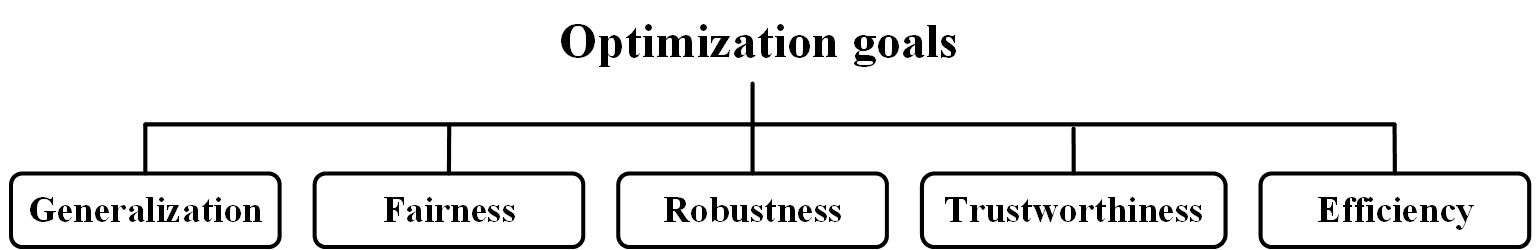}    \vspace{-0.26in}
    \caption{The sub-taxonomy for optimization goals.}
    \label{fig3}
     \vspace{-0.09in}
\end{figure}

On the basis of these principles, the first layer\footnote{The fine-granularity layers are detailed in the succeeding sections.} of our taxonomy is designed as shown in Fig.~{fig2}. This layer consists of six dimensions for data optimization as follows:
\begin{itemize}
    \item Ultimate goals. This dimension refers to the final goal of a data optimization method used in a deep learning task. We divide the optimization goals into five main aspects\footnote{It should be noted that these five aspects are not exhaustive and there are overlaps among them as revealed by the previous literature.}, including generalization, robustness, fairness, trustworthy, and efficiency.
    \item Application scenarios. This dimension refers to the deep learning applications that utilize data optimization. Nine applications are involved, including learning under biased distribution, noisy-label learning, learning with redundant training data, learning with limited training data, model safety, fairness-aware learning, learning under distribution drift, trustworthy learning, and learning for large models.
    \item Data objects. This dimension refers to the objects to optimize in the employed data optimization method. Most studies focus on the raw training data. There are also methods concentrating in other data objects such as hyper-parameters and meta data.
    \item Optimization pipeline. This dimension refers to the common steps for a concrete data optimization method for a deep learning procedure. We divide the pipeline into three common steps, namely, data perception, analysis, and optimizing.
    \item Optimization techniques. This dimension refers to the employed technique paths in data optimization. This study summarizes five main paths, namely, data resampling, data augmentation, data perturbation, data weighting, and dataset pruning. Each path also contains sub-divisions. The introduction for this part is the focus of this survey.
    \item Optimization theories. This dimension refers to the theoretical analysis and exploration for data optimization in deep learning. We divided this dimension into two aspects: formulation and explanation. 
\end{itemize}

Section IV introduces the ultimate goals, application scenarios, and data objects. Sections V, VI, and VII introduces the optimization pipeline, technique, and theory, respectively.

\section{Goals, scenarios, and data objects}
This section introduces ultimate goals, targeted applications, and data objects.

\subsection{Optimization goals}
Fig.~\ref{fig3} describes the sub-taxonomy for the dimension of optimization goals, including generalization, fairness, robustness, trustworthiness, and efficiency.
 
 Generalization is the primary optimization goal in most data optimization techniques, as it is almost the sole goal in most deep learning tasks. According to the generalization theory studied in shallow learning, generalization of a category is highly related to class margin, inter-class distance, and class compactness~\cite{GSI2020}. A larger margin/larger inter-class distance/higher class compactness of a category indicates a better generalization performance on the learned model on the category. The data augmentation strategy that injects noise to training samples is proven to increase the generalization~\cite{LNS2021}. The implicit data augmentation method ISDA~\cite{ISDA2019} actually aims to improve the class compactness\footnote{Some methods such as center loss also aim to increase the class compactness. These methods are considered not data optimization.} of each category. Adaptive margin loss~\cite{MBN2021} also aims to improve the class compactness by perturbing the logits. Fujii et al.~\cite{DAB2022} modified the classical data augmentation method mixup~\cite{MBE2018} by considering the ``between-class distance", which finally increases the inter-class distance. In addition, some studies explore the compiling of an optimal batch in the training process of deep learning~\cite{OBS2016}. The ultimate goal is also the generalization. Nevertheless, the direct goal of the batch compiling may consist of balance, diversity, and others.

As previously stated, fairness is also an important learning goal in many deep learning tasks. To combat unfairness on samples with certain attributes, techniques such as data augmentation~\cite{FMF2021}, perturbation~\cite{FCW2021}, and sample weighting~\cite{FLT2022} have been used in previous literature. Indeed, imbalanced learning also pursues fairness among different categories. A category with a small prior probability, denoted as a minor category, will receive more attention in the employed data optimization. For example, larger weights~\cite{CLB2019}, larger degrees of data perturbation~\cite{LP2022}, or more augmented quantities~\cite{SAA2023} are exerted on minor categories than others.

Adversarial robustness is an essential goal in deep learning tasks that are quite sensitive to model safety. Adversarial training is usually leveraged to improve the adversarial robustness of a model. It can be attributed to a special type of data augmentation. Thus, adversarial training is actually a data optimization technique, which aims to improve the quality of training data such that the models trained on the optimized training data have better adversarial robustness. 

Trustworthiness is a goal that has recently been highly valued. Explainability and calibration are two crucial requirements for the trustworthiness of a deep learning model. Explainability mainly relies on methodologies such as feature attribution and causal reasoning rather than pure data optimization technique. Nevertheless, data optimization is widely used in model calibration. Calibration mainly concerns the trustworthiness of the predicted probability of a probabilistic model~\cite{OWC2015}.  Liu et al.~\cite{TDI2022} introduced margin-aware label smoothing to improve the calibration of trained models. Mukhoti et al.~\cite{CDN2020} leveraged sample weighting to achieve better calibration. 

Efficiency is crucial for real applications as many learning tasks are sensitive to both time complexity and storage. Therefore, how to optimally reduce the redundant training data and remain the diverse important training data deserves further investigation. The time complexity can be significantly reduced after data pruning.

\begin{figure}[t] 
    \centering \includegraphics[width=1\linewidth]{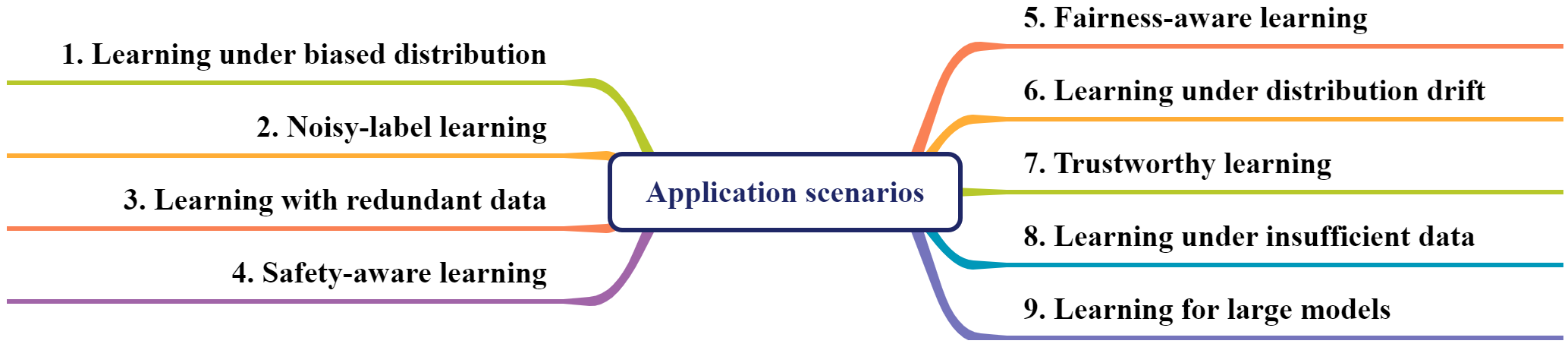}    \vspace{-0.26in}
    \caption{The sub-taxonomy for targeted application scenarios.}
    \label{fig4}
     \vspace{-0.1in}
\end{figure}
\vspace{-0.1in}
\subsection{Application scenarios}\label{section4.2}
Fig.~{fig4} describes the sub-taxonomy for the dimension of targeted application scenarios. The first eight scenarios have been referred to in previous sections, so they are not further introduced in this subsection. 

Learning under insufficient data contains the case that the training data are not as diverse as possible. Data diversity affects the model generalization~\cite{CDD2022}. Dunlap et al.~\cite{DYV2023} utilized large vision and language models to automatically generate visually consistent yet significantly diversified training data. Some studies~\cite{IDI2023,IDA2022} consider that data augmentation is actually a widely-used technique to increase data diversity. These studies develop new data augmentation methods for deep learning tasks. 

Large models, such as large language models~(LLMs), have made remarkable advancements in nearly each AI field. The data quality is crucial for the training or fine-tuning of a large model. Therefore, data optimization techniques also prevail in learning for large models. Yang et al.~\cite{BAL2023} utilized flip operation on the training corpus to balance the two-way translation in language pairs in their building of a large multilingual translation model. Liu et al.~\cite{ATF2020} applied adversarial training in both the pre-training and fine-tuning stages. Results show that the error rate of the trained model is also reduced. 

\begin{figure}[htbp] 
    \centering \includegraphics[width=0.88\linewidth]{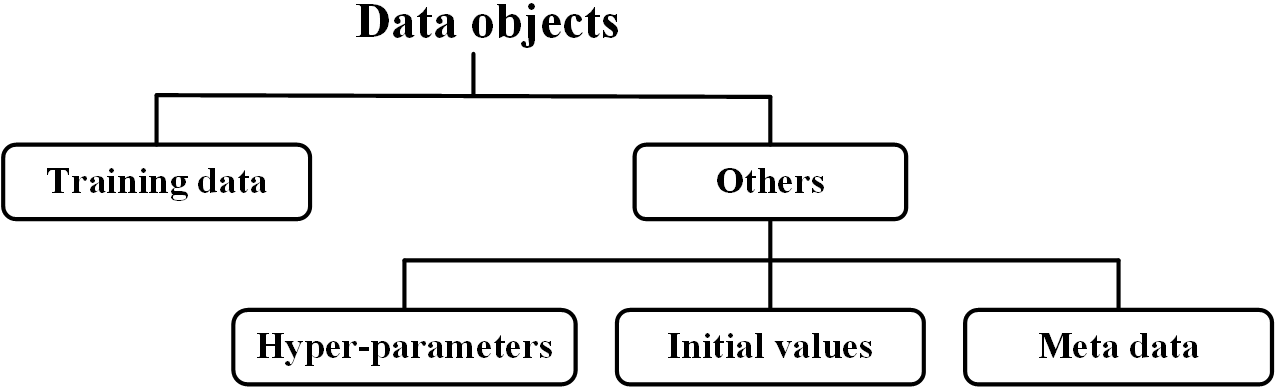}    \vspace{-0.1in}
    \caption{The sub-taxonomy for data objects.}
    \label{fig5}
     \vspace{-0.06in}
\end{figure}

\subsection{Data objects}
\subsubsection{Primary objects}
The primary objects of data optimization for deep learning are the training data. Some studies optimize raw samples, while some others optimize labels. There are also studies focusing on the data transformed by DNNs, e.g., features and logits. In Sections VI-B and VI-C, more details will be presented.

\subsubsection{Other objects}
There are also numerous studies concerning other data objects in deep learning. Fig.~\ref{fig5} lists three other data objects, namely, hyper-parameters, initial values, and meta data. 

Hyper-parameters highly affect the final performance of trained models. They are determined either by grid searching in a pre-defined scope or directly being set as fixed values. Consequently, setting a proper searching scope or fixed initial values is a crucial step in DNN training. 

Network initialization is also important for DNN training. Gaussian distribution-based initialization is the primary choice in most learning tasks. Other effective strategies are also investigated and applied. Glorot and Bengio~\cite{UID2010} adopted a 
scaled uniform distribution for initialization which is called ``Xavier" initialization. He et al.~\cite{DDI2015} proposed a robust initialization method for rectifier nonlinearities, which is called ``Kaiming" initialization.

Meta learning offers a powerful manner to optimize hyper-parameters of independent modules in deep learning. It relies on an unbiased meta dataset. Nevertheless, in most learning tasks there are no independent high-quality meta data and constructing a high-quality unbiased meta dataset is challenging. Su et al.~\cite{SMD2022} conducted a theoretical analysis for the compiling of a high-quality meta dataset from the training set. Four criteria, namely, balance, uncertainty, clean, and diversity, are selected in their proposed compiling method.

There are numerous classical studies for the optimization of the three types of data which are not mentioned in this study. The placing of these studies into data optimization for deep learning may facilitate the further development of the optimization for the three types of data. The focus of this survey is the training data. Therefore, the following parts will be limited in the scope of training data optimization.
\begin{figure}[htbp] 
    \centering \includegraphics[width=0.736\linewidth]{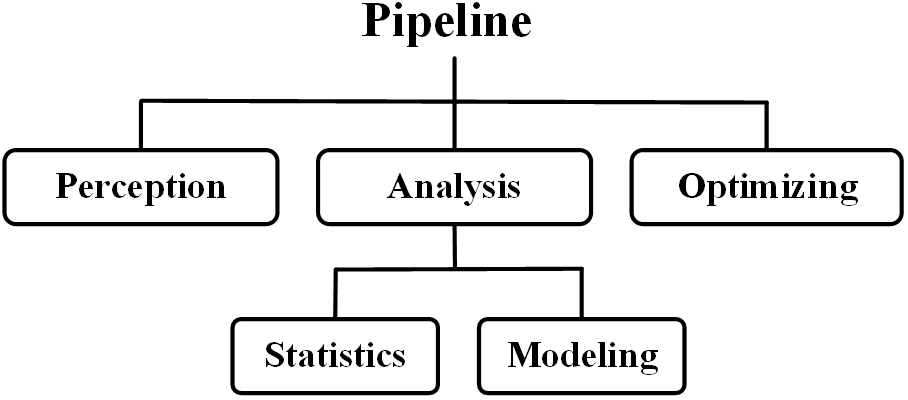}    \vspace{-0.1in}
    \caption{Three main steps in data optimization pipeline.}
    \label{fig6}
     \vspace{-0.1in}
\end{figure}

\begin{figure}[t] 
    \centering \includegraphics[width=1\linewidth]{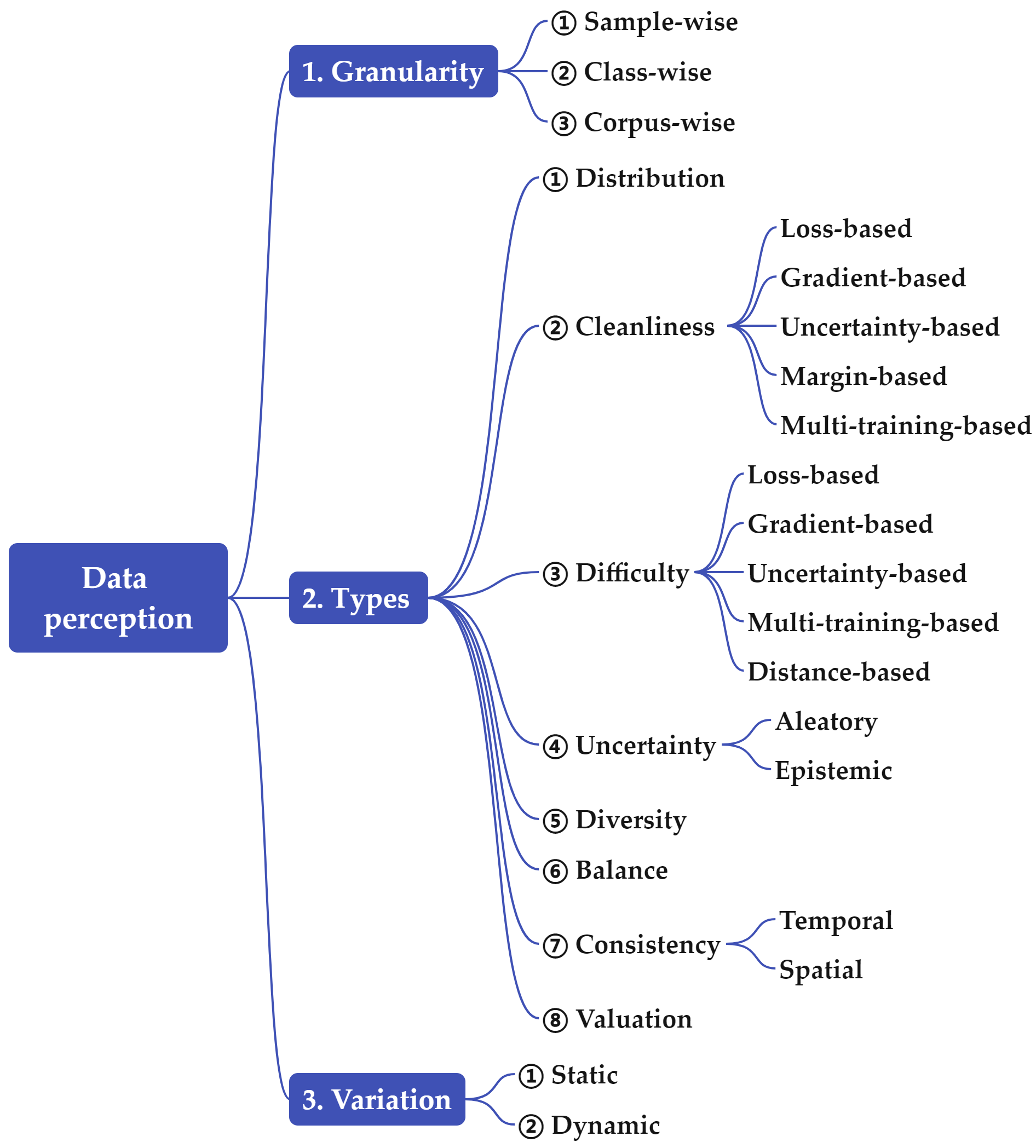}    \vspace{-0.16in}
    \caption{The sub-taxonomy for data perception.}
    \label{fig7}
     \vspace{-0.1in}
\end{figure}

\section{Optimization pipeline}
The pipeline mainly consists of three steps, namely, data perception, analysis, and optimizing, as shown in Fig.~\ref{fig6}. Some notations and symbols are defined as follows. Let $D = \{ x_i, y_i\}_{i=1}^N$ be a set of $N$ training samples, where $x_i$ is the feature and $y_i$ is the label. Let $C$ be the number of categories and $N_c$ be the number of the samples in the $c$th category in $D$. $\pi_c{\rm{ }} = {\rm{ }}N_c/N$ is the proportion of the $c$th category. Let $p_c$ and $p(x|y=c)$ be the prior and the class conditional probability density for the $c$th class, respectively. When there is no ambiguity, $x_i$ represents the feature output by the last feature encoding layer and $u_i$ represents the logit vector output by the Softmax layer for $x_i$. Let $l$ be the loss for $x$. In this study, the cross-entropy loss is assumed and $\Theta$ represents the network parameters.

\subsection{Data perception}
In this study, data perception refers to all possible methods aimed at sensing and diagnosing the training data to capture the intrinsic data characteristics and patterns that affect learning performance. It serves as the first step in the pipeline, and an effective data optimization method cannot work well without accurate perception of the training corpus  

Generally, data perception for training data quantifies the factors related to the true distribution, training data distribution, cleanliness, diversity, etc. We construct a sub-taxonomy for data perception in three dimensions as shown in Fig.~\ref{fig7}. First, in terms of quantifying granularity, there are three levels, namely, sample-wise, category-wise, and corpus-wise. Secondly, in terms of perception types, there are eight divisions, namely, distribution, cleanliness, difficulty, diversity, balance, consistency, neighborhood, and valuation. Thirdly, in terms of quantifying variation, there are two divisions, namely, static and dynamics. Each of the above divisions and partial representative studies are introduced as follows.

\subsubsection{Perception on different granularity levels}
There are three granularity levels, including sample-wise, category-wise, and corpus-wise.

\textbf{Sample-wise data perception}. It denotes that the perceived quantities reflect or influence a sample's positive/negative or trivial/important role in training. For example, most noisy-label learning methods employ sample-wise data perception, e.g., training loss~\cite{OAS2019} and gradient norm~\cite{GHS2019}, to infer the noisy degree of a training sample. 

\textbf{Category-wise data perception}. It denotes that the perceived quantities reflect or influence a category's positive/negative or trivial/important role in training. In category-wise perception, the learning performance of each category is usually monitored to return feedback for the entire scheme~\cite{DDIL2021}. Therefore, the average outcome~(e.g., average loss or precision) is also used to infer a reasonable category-wise weight in the next training epoch~\cite{CDBL2020,CCC2022}. Another popular quantity is the category proportion~($\pi_c$) for imbalanced learning. Some studies~\cite{HCF2023} measure the compactness of a category as it reflects the generalization of the features for a category. Studies on/leveraging category-wise perception are fewer than sample-wise studies. 

\textbf{Corpus-wise data perception}. It denotes that the perceived quantities reflect or influence a training corpus' positive/negative or trivial/important role in training. Limited studies fall into this division. Lin et al.~\cite{MIE2022} used the query score to measure the utility of a training dataset. 

These three levels can be used together to more comprehensively perceive the training data~\cite{CAW2023}.

\subsubsection{Perception on different types}
The eight quantifying types are introduced as follows: 
\begin{itemize}
    \item Distribution. This type aims to quantify the true data distribution for a learning task and the training data distribution. An effective quantification of these two distributions is significantly beneficial for training. Nevertheless, it is nearly impossible to obtain a clear picture of them. Therefore, the true distribution is usually assumed to conform to several some basic assumptions, such as Gaussian distribution for each category~\cite{ISDA2019}. For the training data distribution, some studies~\cite{IFL2022,DAAS2023} apply clustering to deduce the intrinsic structure of the training data. These studies concern the global distribution of a category. Recently, researchers have investigated local distributions of training samples. One typical characteristic is about the neighborhood of each training sample. In deep learning on graphs, the distribution of neighborhood samples with heterogeneous labels negatively impacts the training or prediction for the sample. Wang et al.~\cite{TTI2022} defined a label difference index to quantify the difference between a node and its neighborhood in a graph as follows:

\begin{equation} 
LDI(x_i) = \frac{1}{\sqrt{2}}||p_{x_i}-p_{N_i}||_2,
\end{equation}
where $p_{x_i}$ and $p_{N_i}$ are the category distributions of $x_i$ and its neighborhood $N_i$.
    \item Cleanliness. This type aims to identify the degree of noise in each training sample. This study primarily focuses on label noise, as it garners more attention than sample noise. There are numerous metrics for noise measurement. As illustrated in Fig.~\ref{fig7}, typical measures include loss-based, gradient-based, uncertainty-based, margin-based, and multi-training-based techniques. Samples with large losses, gradient norms, or uncertainties are more likely to be noisy. In margin-based measures, a small margin indicates a high probability of being noise. Huang et al.~\cite{OAS2019} conducted multiple training procedures to identify noisy labels.
   
    \item Difficulty. This type aims to infer the degree of learning difficulty for a training sample or a category. The accurate measurement of learning difficulty for each training sample is of great importance because several deep learning paradigms employ adaptive learning strategies based on the level of learning difficulty. For instance, curriculum learning~\cite{CULE2009} holds the perspective that easy samples should receive more focus in the early training stages, while hard samples should be given more attention in the later stages of training. Some other studies~\cite{FLD2017} hold the opposite perspective that hard samples should be prioritized throughout the training procedure. As shown in Fig.~\ref{fig7}, there are five major manners to measure learning difficulty of samples, namely, loss-based, gradient-based, uncertainty-based, multi-training-based, and distance-based. Obviously, the measures for learning difficulty are quite similar to those for cleanliness. In fact, some studies consider that noisy samples are those quite difficult to learn and divide samples into easy/medium/hard/noisy. Paul et al.~\cite{DLOA2021} proposed the error l2-norm score to measure difficulty. Zhu et al.~\cite{ETL2022} established a formal definition for learning difficulty of samples inspired by the bias-variance trade-off theorem and proposed a new learning difficulty measures. Sorscher et al.~\cite{BNS2022} defined the cosine distance of a sample to its nearest cluster center as the sample's difficulty measure and applied it in sample selection.

    \item Uncertainty. This type contains two sub-types, namely, aleatory uncertainty and epistemic uncertainty~\cite{AROUQ2021}. The former is also called data uncertainty and occurs when training samples are imperfect, e.g., noisy. Therefore, the cleanliness degree can be used as a measure of data uncertainty~\cite{ATOT2021}. Epistemic uncertainty is also called model uncertainty. It appears when the learning strategy is imperfect. Model uncertainty can be calculated based on information entropy of the DNN prediction or the variance of multiple predictions output by a DNN with the dropout trick~\cite{WUD2017}.
    
    \item Diversity. This type aims to identify the diversity of a subset of training samples. The subset is usually a category. The measurement for subset diversity is useful in the design of data augmentation strategy for the subset~\cite{SOBD2019} and data selection~\cite{SMD2022}. Friedman and Dieng~\cite{TVS2023} leveraged the exponential of the Shannon entropy of the eigenvalues of a similarity matrix, namely, vendi score to measure diversity. Salimans et al.~\cite{ITF2016} utilized a pre-trained Inception model to measure diversity called inception score.
    \item Balance. This type aims to measure the balance between/within categories. The balance between categories belongs to global balance, while that within a category belongs to local balance. Global balance can be simply measured by the proportion of the training sample of a category. Nevertheless, our previous study~\cite{RCI2023} reveals that other factors such as variance and distance may also result in serious imbalance. Local balance is relatively difficult to measure. Some studies define local balance as attribute balance~\cite{IFL2022}. 
    \item Consistency. This type aims to identify the consistency of the training dynamics of a training sample along the temporal or spatial dimensions. In the temporal dimension, the variations of the training dynamics between the previous and the current epochs are recorded~\cite{DCM2020}. In the spatial dimension, the differences in the training dynamics between a sample and other samples such as neighbors~\cite{LWN2022} or samples within the same category are recorded. A classical measure called ``forgetting"~\cite{AES2019} quantifies the number of variations in the prediction between adjacent epochs. Singh et al.~\cite{ACF2022} investigated class-wise forgetting. Maini et al.~\cite{CDV2022} utilized forgetting to distinguish among examples that are hard for distinct reasons, such as membership in a rare subpopulation, being mislabeled, or belonging to a complex subpopulation. Wang et al.~\cite{ACSOF2023} provided a comprehensive summary for sample forgetting in learning. Kim et al.~\cite{FSF2021} focused on the dynamics of each sample's latent representation and measured the alignment between the latent distributions.
    \item Valuation. This value is usually measured by the Shapley value, which is a concept from the game theory~\cite{AVF1953}. Ghorbani and Zou firstly introduced Shapley value for data valuation~\cite{DSE2019} as follows:
\begin{equation} 
\phi(x_i) = \sum_{S \in D-x_i}\frac{1}{C_{|D|-1}^{|S|}} [V(S \cup \{x_i\})-V(S)]\label{SV}
\end{equation}
where $V(\cdot)$ is the utility function of a dataset and $S$ is a subset of the training corpus $D$. Their values are different when modeling the clean and the noisy samples. 
 Nevertheless, the calculation for the Shapley value as shown in Eq.~(\ref{SV}) is NP-hard, thereby hindering its use in real applications. Yoon et al.~\cite{DVUR2020} proposed a reinforcement learning-based method for data valuation. Their inferred weights reflect the importance of a sample in learning, which is not equal to the Shapley value. Some other studies~\cite{MIE2022,ELF2022} proposed more practical methods to approximate the Shapley value. Jiang et al.~\cite{OAU2023} established an easy-to-use and unified framework that facilitates researchers and practitioners to apply and compare existing data valuation algorithms. Compared with the aforementioned perception quantities such as cleanliness and difficulty, the Shapley value has a more solid theoretical basis. Therefore, establishing a direct connection among the Shapley value and the quantities listed above deserves further study. 
    
\end{itemize}

This study only lists commonly used measures for data perception. Additionally, there are some other important measures which will be explored in our future work. For example, the neighborhoods for each training sample may vary as the feature encoding network is updated at each epoch. In shallow learning, neighborhood is an important information and many classical methods are based on the utilization of neighborhood. However, previous deep learning methodologies rarely leverage neighborhood information, as the computational complexity for neighborhood identification is high. Some studies adopt a simplified manner to construct the neighborhood. For example, Bahri and Heinrich Jiang~\cite{LAL2021} employed the logit vectors rather than the features to construct neighborhood. The dimension of logit vector is usually much smaller than the feature dimension, so the computational complexity is significantly reduced. It is believable that with the advancement of related computational techniques, neighborhood information will receive increasingly attention in deep learning. Some other important quantities such as problematic score~\cite{DPF2021} and data influence~\cite{TDI2023} in learning, which have large overlaps with the aforementioned quantities, also deserve further exploration.

If the perceived quantities are required to fed into a model, hidden representation for the raw quantities is favored. In meta learning-based sample weighting or perturbation~\cite{LTP2021}, the weights or perturbation vectors for each sample are derived based on the hidden representations of the the perceived quantities. For example, Shu et al.~\cite{MLA2019} extracted the training loss for each sample as the input and fed it into an MLP network containing 100 hidden nodes. In other words, the raw training loss is represented by a 100-dimensional hidden vector. Zhou et al.~\cite{CAW2023} utilized six quantities to perceive the character of a training sample, including loss, margin, gradient norm, entropy of the Sofxmax prediction, class proportion, and average categorical loss. Likewise, these six quantities are also transformed into a 100-dimensional feature vector through an MLP network.

\subsubsection{Static and dynamic perception}
Static perception denotes that the perceived quantities remain unchanged during optimization, whereas dynamic perception denotes that the quantities vary.

In imbalanced learning, category proportion is widely used to quantify a category. It belongs to static perception because this quantity remains unchanged. In noisy-label learning, many studies adopt a two-stage strategy in which the noisy degree of each training sample is measured and the degrees are used in the second training stage~\cite{OAS2019}. In this two-stage strategy, the perception for label noise is static. 

The impact of a training sample usually varies during training. Therefore, compared with static perception, dynamic perception is more prevailing in deep learning tasks. Many studies utilize training dynamics of training samples for the successive sample weighting or perturbation. Such training dynamics also belong to the dynamic perception. The training dynamics including loss, prediction, uncertainty, margin, and neighborhood vary at each epoch. For example, self-paced learning~\cite{SLF2010} determines the weights of each training sample according to their losses in the previous epoch and a varied threshold. Therefore, the weight may also vary in each epoch.

\subsection{Analysis on perceived quantities}
Analysis on perceived quantities contains two manners, namely, statistics and modeling, as shown in Fig.~\ref{fig6}. 
\begin{figure}[t] 
    \centering \includegraphics[width=0.98\linewidth]{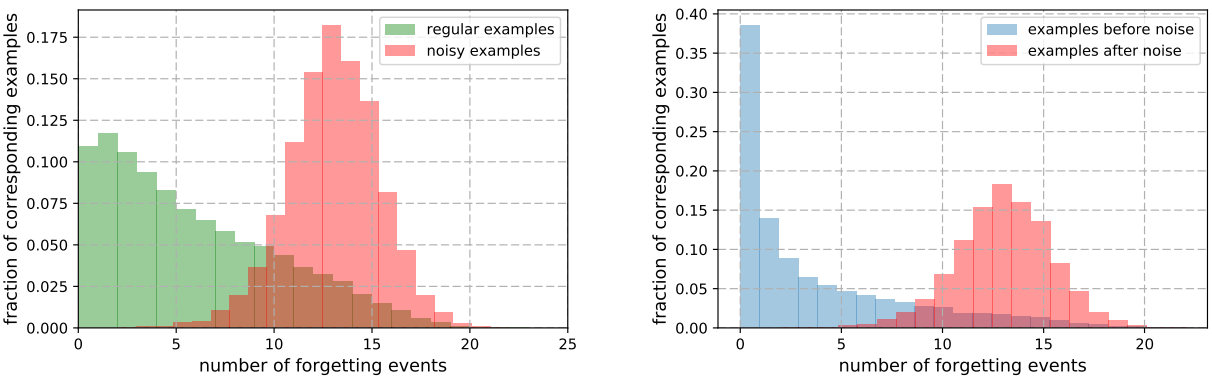}    \vspace{-0.08in}
    \caption{The statistics for the forgetting numbers for training samples~\cite{AES2019}.}
    \label{fig8}
     \vspace{-0.0in}
\end{figure}

\begin{figure}[t] 
    \centering \includegraphics[width=0.98\linewidth]{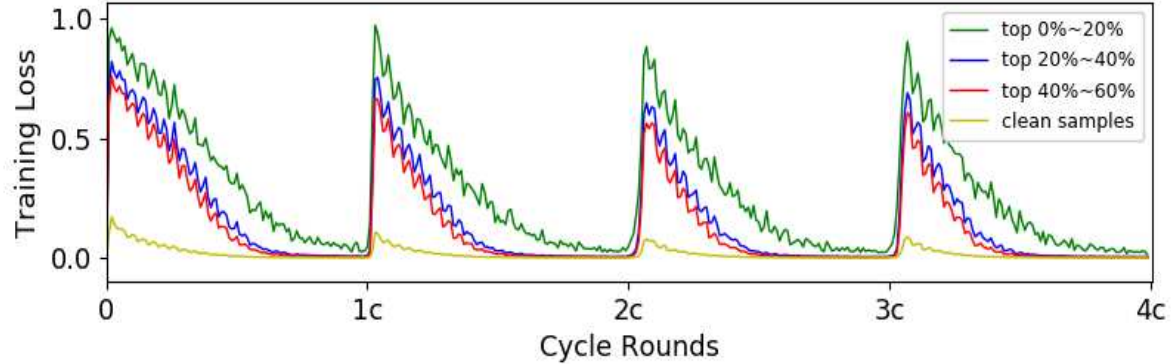}    \vspace{-0.08in}
    \caption{The statistics for loss along the training cycle rounds~\cite{OAS2019}.}
    \label{fig9}
     \vspace{-0.0in}
\end{figure}

\begin{figure}[t] 
    \centering \includegraphics[width=0.98\linewidth]{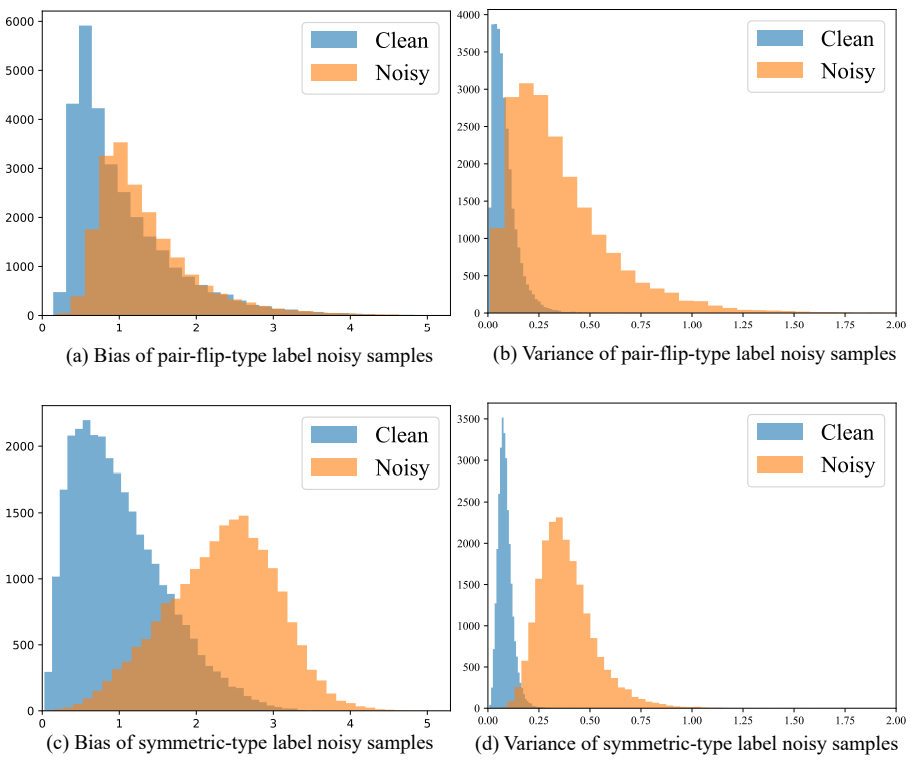}    \vspace{-0.08in}
    \caption{The statistics for the mean and the variance of the loss under the cross validation-based training.~\cite{ETL2022}.}
    \label{fig10}
     \vspace{-0.2in}
\end{figure}

\textbf{Statistical analysis}. Most studies employ this manner for the perceived data quantities. These studies considered only one or two quantities. For example, Toneva et al.~\cite{AES2019} made a statistics for the forgetting numbers of training samples as shown in Fig.~\ref{fig8}. The left figure shows the distributions of forgetting numbers for clean and noisy samples, while the right one shows the distributions of forgetting numbers before and after the noise is added. Distinct difference exists between the distributions of clean and noisy samples. Huang et al.~\cite{OAS2019} proposed a cycle training strategy that the model is trained from overfitting to underfitting cyclically. The epoch-wise loss for each training sample is recorded. Fig.~\ref{fig9} shows the differences between the average losses for the noisy and clean samples. Noisy samples have larger training losses. Therefore, they leveraged the average loss as an indicator for noisy labels. Zhu et al.~\cite{ETL2022} proposed a cross validation-based training strategy. Multiple training losses are also recorded for each training sample. They revealed that the variance of the multiple losses for each sample is also useful in identifying noisy labels as shown in Fig.~\ref{fig10}.


\textbf{Modeling}. This manner refers to the statistical modeling on the perceived quantities for training data. Arazo et al.~\cite{ULN2019} assumed that the traning loss conforms to the following distribution:
\begin{equation} 
p(l|\alpha,\beta) = \frac{\Gamma(\alpha+\beta)}{\Gamma(\alpha)\Gamma(\beta)}l^{\alpha-1}(1-l)^{\beta-1}.
\end{equation}
where $\Gamma(\cdot)$ is the Gamma function; $\alpha$ and $\beta$ are parameters to infer. Their values are different when modeling the clean and the noisy samples. Hu et al.~\cite{MMT2023} leveraged the Weibull mixture distribution to model the memorization-forgetting value of each training sample. The mixture distribution contains two components which suit for the clean and noisy samples, respectively.

Both manners are transparent and thus the entire data optimization approach is explainable. Nevertheless, these two divisions usually rely on appropriate prior distributions about the involved quantities. If the prior distributions are incorrect, the successive optimizing will negatively influence the model training.

\begin{figure*}[htp] 
    \centering \includegraphics[width=0.85\linewidth]{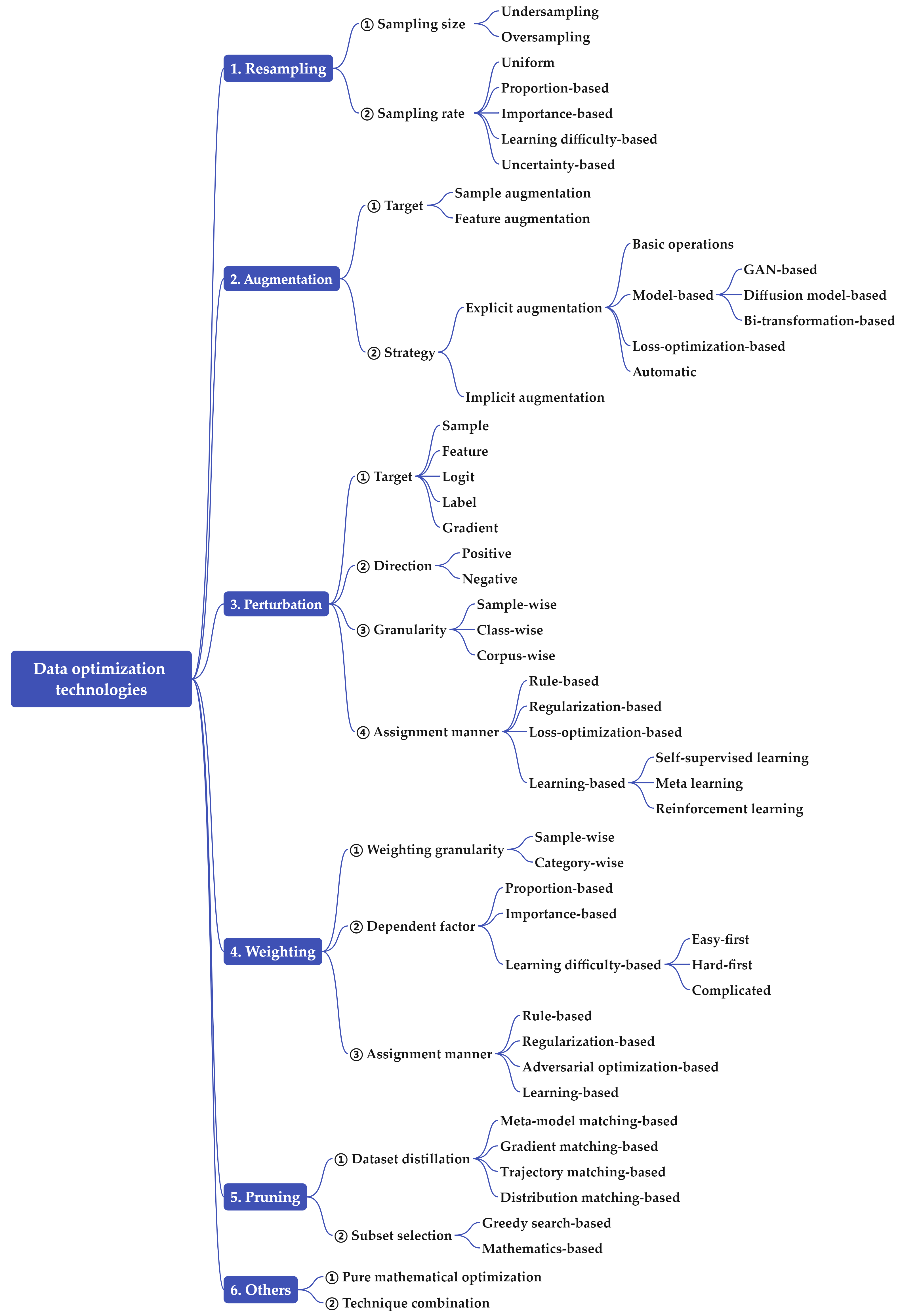}    \vspace{-0.05in}
    \caption{The sub-taxonomy of data optimization techniques.}
    \label{fig11}
     \vspace{-0.0in}
\end{figure*}

\subsection{Optimizing}
The data perception and analysis act as the pre-processing for data operation. This step is the key processing of the entire data optimization pipeline. The successive section will introduce current optimization techniques in detail.

\section{Data optimization techniques}
This section describes the most important dimension for the presented taxonomy, namely, data optimization techniques for deep learning. Fig.~\ref{fig11} presents the sub-taxonomy along this dimension. We summarized six sub-divisions for existing data optimization techniques, including resampling, augmentation, perturbation, weighting, pruning, and others. It is noteworthy that this survey covers numerous technique/methodology divisions and leaves a through comparison for them as our future work. The reason lies in two folds. First, each division has its own merits and defects and their effectiveness have been verified in previous literature, so it is difficulty to judge which one is absolutely the best in universal learning tasks. Second, a thorough theoretical or empirical comparison is not a trivial task.

\subsection{Data resampling}
Data resampling compiles a new training set in which samples are randomly sampled from the raw training set. It is widely used in tasks encountering the issues, including biased distribution~\cite{RRR2019} and redundancy. This study summarizes two split dimensions for this division. The first dimension concerns the size of the sampled datasets, while the second dimension concerns the sampling rates.

In the first dimension, resampling is divided into undersampling and oversampling. undersampling compiles a new training set whose size is smaller than that of the raw training set. Contrarily, oversampling compiles a new training set whose size is larger than that of the raw training set. Both manners are widely used in previous machine learning tasks, including imbalanced learning, bagging, and cost-sensitive learning. Meanwhile, tremendous theoretical studies have been conducted to explain the effectiveness of these two manners in both the statistics and the machine learning communities. Nevertheless, there is currently no consensus on which manner is more effective. Some studies concluded that undersampling should be the primary choice when dealing with imbalanced datasets~\cite{TCI2021}. However, some other studies hold the opposite view~\cite{RRL2023}.

In the second dimension, resampling is divided into uniform, proportion-based, importance-based, learning difficulty-based, and uncertainty-based. They are detailed as follows:
\begin{itemize}
    \item Uniform sampling. This manner is quite intuitive. It treats samples definitely equal regardless of their distributions, location, categories, and training performances. In fact, in nearly all existing deep learning tasks, the batch is constructed by uniformly sampling from the training corpus. Kirsch et al.~\cite{BEA2019} claimed that the independent selection of a batch of samples leads to data inefficiency due to the correlations between samples. Some studies explore alternative sampling strategies. For example, Loshchilov and Hutter~\cite{OBS2016} proposed a rank-based batch selection strategy according to the training loss in previous epochs and samples with large losses have high probabilities to be sampled. Experiments reveal that this new strategy accelerates the training speed by a factor of five.  
    \item Proportion-based sampling. This manner simply assigns the total sampling rate for each category with its proportion~($\pi_c$) in the corpus. It is mainly used in imbalanced learning in which the minor categories are assigned with large sampling rates~\cite{he2009learning}. 
    \item Importance-based sampling. This manner assigns sampling probabilities according to samples' importance. In this study, the definition for importance sampling follows several classical studies~\cite{IPI2000}~\cite{WIT2019}. Given a target distribution $q(x,y)$ and a source distribution on training data $p(x,y)$, the importance (sampling rate) for a training sample $\{x,y\}$ in importance sampling is defined as
    \begin{equation}
        w(x) = \frac{q(x,y)}{p(x,y)}.
        \label{IW1}
    \end{equation}
    As the target distribution is unknown, some studies~\cite{BIS2017} utilize the kernel trick to generate sampling rates.
    In some importance sampling studies, the sampling rates are not based on the probability density ration as presented in Eq.~(\ref{IW1}). For example, Atharopoulos and Fleuret~\cite{NAS2018} took the gradient norms of each training sample as their importance. These methods actually belong to learning difficulty-based sampling.
     \item Learning difficulty-based sampling. This manner assigns sampling rates according to samples' learning difficulties. As summarized in Section V-A2, learning difficulty is usually measured by loss and gradient norm. For instance, Li et al.~\cite{GHS2019} applied the gradient norm of logit vectors as the difficulty measurement. Similar with the study conducted by Atharopoulos and Fleuret~\cite{NAS2018}, Johnson and Guestrin~\cite{TDM2018} proposed the O-SGD sampling method with the following sampling rate:
    \begin{equation}
        w(x,y) = \frac{||\nabla l(x,y)||}{\sum_x ||\nabla l(x,y)||}.
        \label{SW1}
    \end{equation}
   They claimed that this ``importance sampling" can reduce the stochastic gradient’s variance and thus accelerate the training speed. Jiang et al.~\cite{ADL2019} introduced a selective back propagation strategy, in which samples with large losses have relative large probabilities to be selected in back propagation. Gui et al.~\cite{TUD2021} utilized sampling strategy for noisy-label learning. They calculated the sampling weights based on the mean loss of each example along the training process. The training samples with large mean losses are assigned low weights. Liu et al.~\cite{AAM2019} proposed adaptive data sampling, in which the sampling rates of the samples correctly classified in previous epochs are reduced. Xu et al.~\cite{UTR2021} conducted a theoretical analysis and concluded that inverse margin-based sampling may accelerate gradient descent in finite-step optimization by matching weights with the inverse margin.
    \item Uncertainty-based sampling. This manner assigns sampling rates according to samples' uncertainties. It is widely used in active learning, in which a subset of data is sampled for human labeling~\cite{HTM2022,ASOU2021}. Aljuhani et al.~\cite{UAS2022} presented an uncertainty-aware sampling framework for robust histopathology image analysis. The uncertainty is calculated by predictive entropy.
    \end{itemize}
There are also some other sampling manners. For instance, Ting and Brochu~\cite{OSW2018} calculated the sample influence for optimal data sampling. Li and Vasconcelos~\cite{BDR2020} proposed the adversarial sampling to improve OOD detection performance of an image classifier. In their adversarial sampling, the sampling weights are pursued by maximizing the OOD loss. Wang and Wang~\cite{SRF2022} sampled sentences according to their semantic characteristics. Zhang et al.~\cite{UND2019} sampled training data of the majority categories by considering the samples' sensitivities. Samples with low sensitivities may be noisy or safe ones, while those with high sensitivities are borderline ones. Sun et al.~\cite{ASFE2017} explored an automatic scheme for effective data resampling.

\subsection{Data augmentation}
Data augmentation compiles a new training set in which samples~(or features) are generated based on the raw training set or sometimes other relevant sets. It is a powerful tool to improve the generalization capability~\cite{UDA2019,MAD2020} and even adversarial robustness~\cite{DAC2021,DAA2023} of DNNs. Illuminated by related surveys on data augmentation~\cite{ASOD2022,ASOI2019,DAF2022, TSD2021}, two split dimensions are considered, namely, sample/feature and explicit/implicit, as shown in Fig.~\ref{fig11}. 

\subsubsection{Sample/feature augmentation}
In sample augmentation, the new training set consists of generated new samples, while in feature augmentation, the new training set consists of generated new features.

\textbf{Sample augmentation}. This division is subject to data types~(e.g., image, text, or others). For image corpus, augmentation methods adopt noise adding, color transformation, geometric transformation, or other basic operations such as cropping to augment new images~\cite{ASOI2019}. For texts, new samples can be generated by noise adding, paraphrasing, or other basic operations such as word swapping~\cite{DAP2022}. 

\textbf{Feature augmentation}. This division is performed on the feature space, so learning tasks for different data types may utilize the same or similar augmentation strategies. Some intuitive feature augmentation methods include adding noise, interpolating,
or extrapolating~\cite{DAI2017}, which are applicable for general data types, including both image and text data. Li et al.~\cite{ASF2021} revealed that the simply perturbing the feature embedding with Gaussian noise in training leads to comparable domain-generalization performance compared with the SOTA methods. Ye et al.~\cite{AIA2022} proposed novel domain-agnostic augmentation strategies on feature space. Cui et al.~\cite{FSA2020} decomposed features into the class-generic and the class-specific components. They generated samples by combining these two components for minor categories. A classical robust learning paradigm, namely, adversarial training, is actually a feature-wise augmentation strategy when it is run on the feature space~\cite{TDL2018,RDI2021,GAR2023}. 

Some studies augment other data targets such as label and gradient. For example, Lee et al.~\cite{SLA2020} rotated a training image and the rotation angle is also used as supervised information. Elezi et al.~\cite{TLA2018} proposed a transductive label augmentation method to generate labels for unlabeled large set using graph transduction techniques. Some other studies~\cite{AAI2022} investigated gradient augmentation. Compared with sample/feature augmentation, label/gradient augmentation receives quite limited attention.

\subsubsection{Explicit/implicit augmentation}
Explicit augmentation directly generates new samples/features. Meanwhile, implicit augmentation conducts data augmentation only theoretically yet do not generate any new samples/features actually.

\textbf{Explicit augmentation}. According to the employed techniques, existing explicit data augmentation can be divided into basic operation, model-based~(GAN, diffusion model), loss-optimization-based, and automatic augmentation, as described in Fig.~\ref{fig11}. They are introduced as follows:
\begin{itemize}
    \item {Basic operations}. This technique is widely used in practical learning tasks as basic operations conform to human intuitions. The popular deep learning platforms such as pyTorch provide several common basic operations such as cropping, rotation, replacement, masking, cutout, etc. One of the most popular data augmentation method used for shallow learning tasks, namely, SMOTE~\cite{SSM2002} has been utilized in deep learning tasks~\cite{ACDL2022}. Dablain et al.~\cite{DFD2022} designed more sophisticated improvement for SMOTE for deep learning tasks. Among the basic operations, mixup is a simple yet quite effective augmentation manner~\cite{MBE2018,MMB2019}. It generates a new sample with a new label that does not belong to the raw label space.   
    \item {Model-based augmentation}. This technique generates new samples by leveraging independent models. There are three main schemes:
\begin{itemize}
    \item[\ding{172}] GAN-based scheme. Generative adversarial network~(GAN) trains a generative model and a discriminative model simultaneously in a well designed two-player min-max game~\cite{GAN2014}. The trained generative model can be used to generate new samples conforming to the distribution of the involved training data. A large number of variations have been designed in the previous literature~\cite{AROG2023}. Mariani et al.~\cite{BDA2018} proposed balancing GAN for imbalanced learning tasks. Huang et al.~\cite{ACDA2018} developed AugGAN for the data augmentation in cross domain adaptation. Yang et al.~\cite{TTG2022} investigated the GAN-based augmentation for time series.
    \item[\ding{173}] Diffusion model-based scheme. Diffusion models are a new class of generative models and achieve SOTA performance in many applications~\cite{DMA2023}. Xiao et al.~\cite{MDA2023} leveraged a text-to-image stable diffusion model to expand the training set. Dunlap et al.~\cite{DYV2023} utilized large vision and language models to automatically generate natural language descriptions of a dataset's domains and augment the training data via language-guided image editing.

    \item[\ding{174}] Bi-transformation-based scheme. This scheme usually relies on two transformation models. The first model transforms a training sample into a new type of data. The second model transforms the new type of data into a new sample. In natural language processing~(NLP), back-translation is a popular data augmentation technique~\cite{AEE2023}, which translates the raw text sample into new texts in another language and back translates the new texts into a new sample in the same language with the raw sample. Dong et al.~\cite{TLT2017} proposed a new augmentation technique called Image-Text-Image (I2T2I) which integrates text-to-image and image-to-text (image captioning) models. There are also augmentation attempts about Text-Image-Text~(T2I2T)~\cite{SGA2023} and Text-Text-Image~(T2T2I)~\cite{TCG2023}. Theoretically, tri-transformation-based augmentation may be also applicable. We leave it our future work.   
 \end{itemize}
   \item Loss-optimization-based augmentation. This manner generates new sample/features by minimizing or maximizing a defined loss with heuristic or theoretical inspirations. Adversarial training is a typical loss-optimization-based manner. It generates a new sample for $\boldsymbol{x}$ by solving the following optimization problem:
\begin{equation}
\vspace{-0.03in}
{\boldsymbol{x}_{\text{adv}}} =\boldsymbol{x} + \arg \mathop {\max }\limits_{\left\| \boldsymbol{\delta}  \right\| \le \epsilon } \ell(f(\boldsymbol{x} + \boldsymbol{\delta} ),y),
\vspace{-0.03in}
\label{adv}
\end{equation}
where $\boldsymbol{\delta}$ and $\epsilon$ are the perturbation term and bound, respectively. Zhou et al.~\cite{CAW2023} proposed anti-adversaries by solving the following optimization problem:
\begin{equation}
\vspace{-0.03in}
{\boldsymbol{x}_{\text{anti-adv}}} =\boldsymbol{x} + \arg \mathop {\min }\limits_{\left\| \boldsymbol{\delta}  \right\| \le \epsilon } \ell(f(\boldsymbol{x} + \boldsymbol{\delta} ),y).
\vspace{-0.03in}
\label{adv}
\end{equation}
Pagliardini et al.~\cite{IGV2022} obtained new samples by maximizing an uncertainty-based loss.
   \item Automatic augmentation. This manner investigates automated data augmentation techniques~\cite{RPA2019} based on meta learning~\cite{MLAI2022} or reinforcement learning~\cite{ADAV2020}. Nishi et al.~\cite{ASFL2021} proposed new automated data augmentation method and validated its usefulness in noisy-label learning. Some studies focus on differentiable automatic data augmentation which can dramatically reduces the computational complexity of existing methods~\cite{DADA2020}.
\end{itemize}

\textbf{Implicit augmentation}. Wang et al.~\cite{ISDA2019} proposed the first implicit augmentation method called ISDA. It establishes a Gaussian distribution $\mathcal{N}(\boldsymbol{\mu}_{y},\Sigma_{y})$ for each category. New samples can be generated~(i.e., sampled) from its corresponding Gaussian distribution. An upper bound of the loss with augmented samples can then be derived when the number of generated samples for each training sample approaches to infinity. Finally, the upper bound of the loss is used for the final training loss. There are several variations for ISDA, such as IRDA~\cite{IRDA2022} and ICDA~\cite{ICDA2023}. Li et al.~\cite{OFN2021} also proposed an implicit data augmentation approach mainly based on heuristic inspirations.

Explicit augmentation is the primary choice in data augmentation tasks. Nevertheless, implicit augmentation is more efficient than explicit augmentation as it does not actually generate new samples or features. There are also theoretical studies about data augmentation. A mainstream perspective is that data augmentation performs regularization in training~\cite{IRR2019,MLL2019,AOAS2022,TTT2023}. Chen et al.~\cite{AGF2020} conducted a probabilistic analysis and concluded that data augmentation can result in variance reduction and thus prevent overfitting according to the bias-variance theory. In fact, the regularizer in machine learning also reduces model variance.  

\subsection{Data perturbation}
Given a datum ${x}$~($x$ can be the raw sample, feature, logit, label, or others), data perturbation will generate a perturbation $\triangle x$ such that $x'=x+\triangle x$ can replace $x$ or be used as a new datum. Therefore, some data augmentation methods, such as adversarial perturbation, cropping, and masking, can also be viewed as data perturbation. In our previous work~\cite{CL2022}, we constructed a taxonomy for compensation learning, which is actually learning with perturbation. This study follows our previous taxonomy in \cite{CL2022} with slight improvements. The sub-taxonomy for data perturbation is presented in Fig.~\ref{fig11}. Four split dimensions are considered, namely, target, direction, granularity, and assignment manner.

\subsubsection{Perturbation target}
The perturbation targets can be raw sample, feature, logit vector, label, and gradient. 
\begin{itemize}
    \item Sample perturbation. This division adds the perturbation directly to the raw samples. The basic operations in data augmentation can be placed into this division. For instance, noise addition and masking used in image classification actually exert a small perturbation on the raw image. 
    \item Feature perturbation. This division adds the perturbation on the hidden features. Jeddi et al.~\cite{LAE2020} perturbed the feature space at each layer to increase uncertainty in the network. Their perturbation conforms to the Gaussian distribution. Shu et al.~\cite{ERT2021} designed a single network layer that can generate worst-case feature perturbations during training to improve the robustness of DNNs.
    \item Logit perturbation. This division adds the perturbation on the logit vectors in the involved DNNs. Li et al.~\cite{LP2022} analyzed several classical learning methods such as logit adjustment~\cite{LLV2021}, LDAM~\cite{LID2019}, and ISDA~\cite{ISDA2019} in a unified logit perturbation viewpoint. They proposed a new logit perturbation method and extended it to the multi-label learning tasks~\cite{CLP2023}. 
    \item Label perturbation. This division adds the perturbation on either the ground-truth label of the predicted label. One classical learning skill, namely, label smoothing~\cite{RTIA2016}, is a kind of label perturbation method. Let $C$ be the number of categories and $\lambda$ be a hyper-parameter. Label smoothing perturbs the label y~(one-hot type) with the following perturbation $\triangle y = \lambda(\frac{I}{C}-y)$, where $I$ is a $C$-dimensional vector and its each element is 1. A large number of variations have been proposed for label smoothing~\cite{ARVL2019,FLS2021,DDIL2021}.
    \item Gradient perturbation. This division adds the perturbation directly on gradient. Studies on gradient perturbation are few. Orvieto et al.\cite{ANI2022} proposed a gradient perturbation method and verified its effectiveness both theoretically and empirically.
\end{itemize}

There are also studies~\cite{AWP2020} which perturb other data such as network weights in training, which is not the focus of this study. Wang et al.~\cite{RLWP2020} proposed a reward perturbation method for noisy reinforcement learning.

\subsubsection{Perturbation direction}
Data perturbation will either increase or decrease the loss values of training samples in the learning process. Based on whether the loss increases or decreases, existing methods can be categorized as positive or negative augmentations. 

\textbf{Positive perturbation}. It increases the training losses of perturbed training samples. Obviously, adversarial perturbation belongs to positive perturbation, as it maximizes the training loss with the adversarial perturbations. ISDA~\cite{ISDA2019} also belongs to positive perturbation as it adds positive real numbers to the denominator of the Softmax function. 

\textbf{Negative perturbation}. It reduces the training losses. Anti-adversarial perturbation~\cite{CAW2023} belongs to negative perturbation, as it minimizes the training loss with the adversarial perturbations. Bootstrapping~\cite{TDNN2015} is a typical robust loss based on label perturbation. It also belongs to negative perturbation as its perturbation is $\triangle y = \lambda(p-y)$, where $p$ is the prediction of the current trained model.

Some methods increase the losses of some samples and decrease those of others simultaneously. For instance, the  losses of noisy-label training samples may be reduced, while those of clean samples may be increased in label smoothing. Li et al.~\cite{LP2022} proposed a conjecture for the relationship between loss increment/decrement and data augmentation.

\subsubsection{Perturbation granularity}
According to perturbation granularity, existing methods can be divided into sample-wise, class-wise, and corpus-wise.

 \begin{itemize}
     \item Sample-wise perturbation. In this division, each training sample has its own perturbation and different samples usually have distinct perturbations. The aforementioned Bootstrapping and adversarial perturbation all belong to this division. The random cropping and masking also belong to this division. 

     \item Class-wise perturbation. In this division, all the training samples in a category share the same perturbation, and different categories usually have distinct perturbations. Benz et al.~\cite{UATW2021} proposed a class-wise adversarial perturbation method. Wang et al.~\cite{BLV2023} introduced class-wise logit perturbation for the training in semantic segmentation. Label smoothing also belongs to this division.
     \item Corpus-wise perturbation. In this division, all the training samples in the training corpus share only one perturbation. Shafahi et al.~\cite{UAT2017} pursued the universal adversarial perturbation for all the training samples, which has proven to be effective in various applications~\cite{UAPA2020}. Wu et al.~\cite{DBL2020} proposed a corpus-wise logit perturbation method for multi-label learning tasks.
 \end{itemize}

\subsubsection{Assignment manner}
The perturbation variables should be assigned before or during training. As presented in Fig.~\ref{fig11}, there are four typical assignment manners to determine the perturbations.

\textbf{Rule-based assignment}. In this manner, the perturbation is assigned according to pre-fixed rules. These rules are usually based on prior knowledge or statistical inspirations. In both label smoothing and Booststrapping loss, the label perturbation is determined according to manually defined formulas. In text classification, word replacement and random masking also obey rules. 

\textbf{Regularization-based assignment}. In this manner, a regularizer for the perturbation is usually added in the total loss. Take the logit perturbation as an example. A loss function with regularization for logit perturbation can be defined as follows:
\begin{equation} 
\mathcal{L} = \sum_i l(\mathcal{S}(v_i+\triangle v_i),y_i) + \lambda Reg(\triangle v_i).
\end{equation}
where $\mathcal{S}$ is the Softmax function, $v_i$ is the logit vector for $x_i$, $\triangle v_i$ is the perturbation vector for $v_i$, and $Reg(\cdot)$ is the regularizer. Zhou et al.~\cite{TAP2018} introduced a
novel perturbation way for adversarial examples by leveraging smoothing regularization on adversarial perturbations. Wei et al.~\cite{SAP2019} took the notation that adversarial perturbations are temporally sparse for videos and then proposed a sparse-regularized adversarial perturbation method. Zhu et al.~\cite{IAN2023} proposed a Bayesian neural network with non-zero mean of Gaussian noise. The mean is actually a feature perturbation and inferred with $l_2$ regularization.

\textbf{Loss-optimization-based assignment}. This division is similar to the loss-optimization-based augmentation introduced in Section VI-B2. A new loss containing the perturbations is defined and the perturbation is pursued by optimizing the loss. In the optimization procedure, only the perturbations are the variables to be optimized, while the model parameters are fixed.

\textbf{Learning-based assignment}. In this manner, the perturbation is assigned by leveraging a learning method. Three learning paradigms are usually applied, including self-supervised learning, meta learning, and reinforcement learning. 
\begin{itemize}
    \item Self-supervised learning. This paradigm leverages self-supervised learning methodologies such as contrastive learning~\cite{ASFFC2020} to pursue the perturbations. Naseer et al.~\cite{ASAF2020} constructed a self-supervised perturbation framework to optimize the feature distortion for a training image. Zhang et al.~\cite{GAS2022} proposed a generative adversarial network-based self-supervised method to generate EEG signals.
    \item Meta learning. This paradigm leverages meta-learning methodologies to pursue the perturbations using an additional meta dataset. It assumes that the perturbation $\triangle{x}$ (or $\triangle{y}$) for a training sample $x$ (or its label $y$) is determined by the representation of $x$ or factors such as training dynamics for $x$, which is described as follows:
    \begin{equation} 
       \triangle{x} = g(x,\eta(x)),
    \end{equation}
where $g(\cdot)$ can be a black-box neural network such as MLP; $\eta(x)$ represents the training dynamics for $x$. Li et al.~\cite{MMS2021} applied meta learning to directly optimize the covariant matrix used in ISDA, which is used to calculate the logit perturbation. Qiao and Peng~\cite{UMG2021} utilized the meta learning to learn an independent DNN for both features and label perturbation.

\item Reinforcement learning. This paradigm leverages reinforcement learning to pursue the perturbations without relying on additional data. Many data augmentation methods~\cite{ALAP2019,ALDA2019}, which also belong to data perturbation, are based on reinforcement learning. Giovanni et al.~\cite{DRV2020} leveraged deep reinforcement learning to automatically generate realistic attack samples that can evade detection and train producing hardened models. Lin et al.~\cite{ARI2022} formulated the perturbation generation as a Markov decision process and optimized it by reinforcement learning to generate perturbed instructions sequentially.
\end{itemize}
Given a learning task, it is difficulty to directly judge which assignment manner is the most appropriate without a thorough and comprehensive understanding for the task. Each assignment manner has its own merits and defects.

\subsection{Data weighting}
Data weighting assigns a weight for each training sample in loss calculation. It is among the most popular data optimization techniques in many learning scenarios, including fraud
detection~\cite{CCF2018}, portfolio selection~\cite{CPS2022}, medical diagnosis~\cite{ITW2020}, and fairness-aware learning~\cite{FLT2022,FIG2023}. Three dividing dimensions are considered, namely, granularity, dependent factor, and assignment manner for weights. 

\subsubsection{Weighting granularity}
According to the granularity of weights, existing  weighting methods can be divided into sample-wise and category-wise. Noisy-label learning usually adopts sample-wise weighting methods~\cite{DIF2021,SAL2020}, while imbalanced learning usually adopts category-wise ones~\cite{DAAF2021,CLB2019,DWBC2022}. Data weighting is also widely used in standard learning~\cite{JTT2021,GIA2021}, which are usually sample-wise.

\subsubsection{Dependent factor}
Dependent factor in this study denotes the factors that are leveraged to calculate the sample weights. Similar with the resampling introduced in Section VI-A, three factor types are usually considered, namely, category proportion, importance, and learning difficulty. As these concepts are introduced in Section VI-A and quite similar procedures are adopted, these factors are not detailed in this part. There are an increasing number of studies employing learning difficulty-based weighting. They can be further summarized according to which samples are learned first.

 As samples with larger weights than others can be considered as having priority in training, learning difficulty-based weighting contains three basic folds, namely, easy-first, hard-first, and complicated.
\begin{itemize}
    \item Easy-first. Easy samples are given higher
weights than hard ones in this fold. There are a huge number of easy-first weighting methods, which mainly belong to two paradigms: curriculum learning~\cite{CULE2009} and self-paced learning~\cite{SLF2010}. These two paradigms assign larger weights to easy samples during the early training stage and gradually increase the weights of hard samples. Numerous studies have been conducted on the design of the weighting formulas~\cite{ESF2014,SPLW2014,ASML2015}. Easy-first weighting is usually used in noisy-label learning. Extensive experiments on curriculum learning indicate that it mainly takes effects on noisy-label learning tasks~\cite{CLAW2021}.

    \item Hard-first. Hard samples have higher weights than easy ones in this fold. Focal loss is a typical hard-first strategy~\cite{FLD2017}. Zhang et al.~\cite{GIA2021} also assigned large weights on hard samples. Santiagoa et al.~\cite{LTD2021} utilized the gradient norm to measure learning difficulty and exerted large weights on samples with large gradient norms.
    \item Complicated. In some weighting methods, the easy-first or the hard-first is combined with other weighting inspirations. In Balanced CL~\cite{CLW2020}(should be replaced), on the basis of the easy-first mode, the selection of samples has to be balanced under certain constraints to ensure diversity across image regions or categories. Therefore, Balanced CL adopts the complicated mode.
\end{itemize}

Besides the three general ways, Zhou et al.~\cite{WSS2023} revealed some other priority types including both-ends-first and varied manners during training. There also other dependent factors such as misclassified cost and those reflecting other concerns such as fairness and confidence~\cite{MLT2019,CLE2021}.

 \subsubsection{Assignment manner}
Generally, there are four manners to assign weights for training samples as shown in Fig.~\ref{fig11}.

\textbf{Rule-based assignment}. This manner determines the sample weights according to theoretical or heuristic rules. For example, many methods assume that the category proportion is the prior probability. Consequently, the inverse of the category proportion is used as the weight based on the Bayesian rule. Cui et al.~\cite{CLB2019} established a theoretical framework for weight calculation based on the effective number theory in computation geometry. The classical Focal loss~\cite{FLD2017} heuristic defines the weight using $w = (1-p)^{\gamma}$, where $p$ is the prediction on the ground-truth label and $\gamma$ is a hyper-parameter. Han et al.~\cite{AIN2022} defined an uncertainty-based weighting manner for the two random samples in mixup. Importance weighting~\cite{CWN2015} is also placed in this division.

\textbf{Regularization-based assignment}. This method defines a new loss function which contains a weighted loss and a regularizer~($Reg(W)$) on the weights as follows:
\begin{equation} 
\mathcal{L} = \frac{1}{N}\sum_{i=1}^N w_i l(f(x_i),y_i) + \lambda Reg(W),
\end{equation}
where $W=\{w_1,\cdots,w_N\}^T$ is the vector of sample weights. The classical self-paced learning, which mimics the mechanism of human learning from easy to hard gradually, is actually the regularization method defined as $Reg(W) = -|W|_1$~($w_i \in \{0,1\}$)~\cite{SLF2010}. Fan et al.~\cite{SLAI2017} presented a new group of self-paced regularizers deduced from robust loss functions and further analyzed the relation between the presented regularizer-based optimization and half-quadratic optimization. 

\textbf{Adversarial optimization-based assignment.} This manner pursues the sample weights by optimizing a defined objective function, which is similar to the pursing of the adversarial perturbation. For instance, Gu et al.~\cite{ARF2021} adversarially learned the weights of source domain samples to align the source and target domain distributions by maximizing the Wasserstein distance. Yi et al.~\cite{RAS2021} defined a  maximal expected loss and obtained a simple and interpretable closed-form solution for samples' weights: larger weights should be given to augmented samples with large loss values. 

\textbf{Learning-based assignment.} Similar with that in data perturbation, learning-based assignment also usually applies meta learning or reinforcement learning to infer the sample weights. Ren et al.~\cite{LTRE2018} firstly introduced meta learning for sample weighting in imbalanced learning and noisy-label learning. Shu et al.~\cite{MLA2019} utilized an MLP network to model the relationship between samples' characters and their weights, and then trained the network using meta learning. Zhao et al.~\cite{APFF2023} further proposed a probabilistic formulation for meta learning-based weighting. Trung et al.~\cite{UDAF2022} leveraged meta learning to train a neural network-based self-paced learning for unsupervised domain adaption. Wei et al.~\cite{MSLFC2021} also combined meta learning and self-paced network to automatically generate a weighting scheme from data for cross-modal matching. Li et al.~\cite{MRF2023} proposed meta learning-based weighting for pseudo-labeled target samples in unsupervised domain adaptation. Meta learning requires additional meta data, whereas reinforcement learning does not require additional data. Zhou et al.~\cite{MSDA2021} leveraged an augmentation policy network which takes a transformation and the corresponding augmented image as inputs to generate the loss weight of an augmented. Ge et al.~\cite{ADDF2023} used a delicately designed controller network to generate sample weights and combined the weights with the loss of each input data to train a recommendation system. The controller network is optimized by reinforcement learning.
  
Weights assignment can also be divided into static and dynamic. There are a few methods adopting static weighting~\cite{CLB2019}, whereas most methods adopting dynamic. Fang et al.~\cite{RIW2020} proposed dynamic importance weighting to train the models.

\subsection{Data pruning}
Data pruning is contrary to data augmentation. In this study, it is divided into dataset distillation and subset selection.

\subsubsection{Dataset distillation}
Dataset distillation is firstly proposed by Wang et al.~\cite{DDist2018} and it aims to synthesize a small typical training set from substantial data~\cite{ACSOD2023}. The synthesized dataset replaces the given dataset for efficient and accurate data-usage for the learning task. Following the division established by Sachdeva and McAuley~\cite{DDA2023}, existing data distillation methods can be placed in four folds.

\textbf{Meta-model matching-based strategy.} This strategy is firstly proposed by Wang et al.~\cite{DDist2018}. It performs an inner-loop optimization for a temporal optimal model based on the synthesized set and an outer-loop optimization for a temporal subset~(i.e., the synthesized set) by turns. Some recent studies discussed its drawbacks such as the ineffectiveness of the TBPTT-based optimization~\cite{DCWG2021} and proposed new solutions such as momentum-based optimizers~\cite{RTD2022}. Loo et al.~\cite{EDD2022} utilized the light-weight empirical neural network Gaussian process kernel for the inner-loop optimization and a new loss for outer-loop optimization. Zhou et al.~\cite{DDU2022} combined feature extractor in the distillation procedure.

\textbf{Gradient matching-based strategy.} This strategy~\cite{DCWG2021, DCWD2021} does not require to perform the inner-loop optimization as used in the meta-model matching-based strategy. Therefore, it is more efficient than the meta-model matching-based strategy. Numerous approaches have been proposed along this division. Kim et al.~\cite{DCV2022} further utilized spatial redundancy removing to accelerate the optimization process and gradients matching on the original dataset.

\textbf{Trajectory matching-based strategy.} This strategy performs distillation by matching the training trajectories of models trained on the original and the pursued datasets~\cite{DDB2022}. Cui et al.~\cite{SUD2022} proposed a memory-efficient method which is avaliable for large datasets.

\textbf{Distribution matching-based strategy.} This strategy performs the distillation by directly matching the distribution of the original dataset and the pursued dataset~\cite{DCWD2023}. Wang et al.~\cite{CLT2022} constructed a bilevel optimization strategy to jointly optimize a single encoder and summarize data. 
 
There are some solutions~\cite{PBC2022,SIT2022,BSL2022,ILT2023,RTD2022,DDVF2022} which take alternative technical strategies. Zhou et al.~\cite{PBC2022} introduced reinforcement learning to solve the bi-level optimization in data distillation. Zhao and Bilen~\cite{SIT2022} learned a series of low-dimensional codes to generate highly informative images through the GAN generator. 

\subsubsection{Subset selection}
Different from dataset distillation that generates a new training set, subset selection aims to select the most useful samples from the original training set~\cite{ASODO2022}. It does not generate new samples and can be used in any training stages that require to select samples from the original training set. In Fig.~\ref{fig11}, there are two divisions, including greedy search-based and mathematics-based. 

In the greedy search-based strategy, the utility of each training sample is measured, and the subset is searched based on the utility rankings. According to the employed measures, existing methods can be divided into four categories, including difficulty-based, influence-based, value-based, and confidence-based. Meding et al.~\cite{TOI2022} utilized the misclassified rate by multiple classifiers as the learning difficulty of a training sample to select samples. Feldman and Zhang~\cite{WNN2020} defined an influence score and a memorization score to measure the usefulness of a training sample. Samples with low influence and memorization scores are redundant and can be deleted. Birodkar et al.~\cite{SRI2019} employed clustering to select most valuable samples which are close to the cluster centers and delete the rest redundant ones. Northcutt et al.~\cite{LWC2017} leveraged the confidence score to prune training samples. There are also many studies combining the measures and active learning to select samples~\cite{LFLD2019}.

Different from the greedy search strategy, some other methods seek a global optimal subset according to a mathematical approach. Yang et al.~\cite{TSL2023} proposed a scalable framework to iteratively extract multiple mini-batch coresets from larger random subsets of training data by solving a submodular cover problem. Mirzasoleiman et al.~\cite{CFD2020} defined a monotonic function for coreset selection and proposed a generic algorithm with approximately linear complexity.

\subsection{Other typical techniques}
This study lists two representative technical paths, including pure mathematical optimization and the combination of more than one aforementioned methods described in Sections VI-A to VI-E.

\subsubsection{Pure mathematical optimization}
This division refers to the manners that perform data optimization via a pure mathematical optimization procedure in the above-mentioned divisions.

The first typical scenario for pure mathematical optimization is the construction of a small-size yet high-quality dataset from the original training set. The tasks involving batch construction, meta data compiling in meta learning, or dataset distillation usually adopt mathematical optimization. Liu et al.~\cite{DPA2021} constructed a set variance diversity-based objective function for data augmentation and pursued the selection for a set of augmented samples via the maximization of the objective function in batch construction. Joseph et al.~\cite{SBS2019} proposed a submodular optimization-based method to construct a mini-batch in DNN training. Significant improvements in convergence and accuracy with their constructed mini-batches have been observed. Su et al.~\cite{SMD2022} established an objective function for meta data compiling. The objective consists of four criterion, including cleanliness, diversity, balance, and informative. As introduced in Section VI-E, data pruning is usually performed based on pure mathematical optimization.  

The second typical scenario is the regularized sample weighting or perturbation. The details are described in Sections VI-C4 and VI-D3. For instance, Li et al.~\cite{RVS2020} devised a new objective function for the label perturbation strength, which can also reduce the Bayes error rate during training. Meister et al.~\cite{GER2020} constructed a general form of regularization that can derive a series of label perturbation methods.

The third typical scenario is the constrained optimization, which embeds prior knowledge or conditions in data weighting, perturbation, or pruning into the constraints. For instance, Chai et al.~\cite{FWA2022} defined an optimization objective function with the constraints that each demographic group should have equal total weights in fairness-aware learning. The adversarial perturbation of multi-label learning is usually attained by solving constrained optimization problems~\cite{MAP2018,TAAT2021,EMA2023}. Hu et al.~\cite{TAAT2021} developed a novel loss for multi-label top-$k$ attack with the constraints that considers top-$k$ ranking relation among labels.

\subsubsection{Technique combination}
Indeed, many learning algorithms do not employ a single data optimization technique. Instead, they combine different data optimization techniques. The following lists a few combination examples. 

In data augmentation, many methods choose to generate samples in the first step and resample or reweight the samples in the second step. For instance, Cao et al.~\cite{MEB2023} dealt with grammatical error correction by using a data augmentation method during training and a data weighting method to automatically balance the importance of each kind of augmented samples. Liu et al.~\cite{CDAF2021} generated new source phrases from a masked language model then sampled an aligned counterfactual target phrase for neural machine translation. Zang et al.~\cite{FFA2021} combined data augmentation and resampling for a long-tailed learning task.

In data perturbation, different directions/granularity levels are usually combined in the same method. For example, adversarial perturbation belongs to the positive direction, while anti-adversarial perturbation belongs to the negative one. Zhao et al.~\cite{ALA2022} considered both category-wise and sample-wise factors to define the logit perturbation for imbalanced learning. Zhou et al.~\cite{CAW2023} combined both adversarial and anti-adversarial perturbations and theoretically revealed the superiority of the combination than the adversarial perturbation only. 

In data weighting, numerous methods combine it with data augmentation. Han et al.~\cite{AIN2022} combined uncertainty-based weighting and the classical augmentation method mixup. Chen et al.~\cite{IRDA2022} combined effective number-based weighting and logit perturbation for long-tail learning tasks. In addition, some methods combine different granularity levels or different priority models. For example, Focal loss~\cite{FLD2017} employs both category-wise and sample-wise weight coefficients for each sample.

\section{Data optimization theories}
There are a large amount of studies focusing on the theoretical aspects of data optimization. It is quite challenging to arrange existing theoretical studies into a clear roadmap. This study summarizes existing studies in the following two dimensions, including formalization and explanation.

\subsection{Formalization}
In order to theoretically analyze and understand the data optimization methods, it is essential to establish mathematical formulations. Statistical modeling is the primary tool for their formalization~\cite{WDR2015, AEA2019, DEA2023}. Basic assumptions are usually relied on. The most widely used assumptions for the statistical modeling include the following.

\begin{itemize}
    \item Gaussian distribution assumption. Many studies~\cite{DID2021,FFA2022,LVR2022, BMF2022} assume that data in each category conforms to a Gaussian distribution, which simplifies computation and inference compared to other complicated distributions~\cite{GDB2022}.
    \item Equal class CPD assumption. In many learning studies~\cite{RTV2020,SLI2022} excepting those for distribution drift, the class-conditional probability densities~(CPD) of the training and testing sets are assumed to be identical. 
    \item Uniform distribution assumption. In many studies\cite{LID2019,AOT2023}, the distribution over categories in the testing set is assumed to be uniform. Some studies implicitly use this assumption by using modified losses such as the balanced accuracy or balanced test error~\cite{IDL2019,LLV2021}, even if the category proportions in the test corpus are not identical.
    \item Linear boundary assumption. In many studies~\cite{IIW2022,UTR2021}, the decision boundary of the involved classifier is assumed to be linear. The decision boundary between two categories under the cross-entropy loss is linear.
\end{itemize}

Based on these assumptions, the data optimization problems are usually formalized into probabilistic, constrained optimization, or regularization-based problems. For example, Xu et al.~\cite{ATA2022} investigated importance weighting for covariate-shift generalization based on probabilistic analysis. Chen et al.~\cite{ZLA2022} defined the classification accuracy based on posterior probability for zero-shot learning. Qraitem et al.~\cite{BMA2023} formalized a constrained linear program problem to investigate the effect of data resampling. Roh et al.~\cite{SSF2021} formulated a combinatorial optimization problem for the unbiased selection of samples in the presence of data corruption. In classical weighting paradigm such as SPL, data weighting is directly formalized in the optimization object consisting of the weighted loss and a regularizer. Zhang et al.~\cite{BCD2023} defined a re-weighted score function consisting of weighted loss and a sparsity regularization for causal discovery. 

Jiang et al.~\cite{ADV2019} proposed a new adversarial perturbation generation method by adding a diversity-based regularization which measures the diversity of candidates. Hounie et al.~\cite{ADA2023} proposed a constrained learning problem for automatic data augmentation by combining conventional training loss and the constraints for invariance risk. Blum and Stangl~\cite{RFB2019} investigated the utility of fairness constraints in fair machine learning.

\subsection{Explanation}
Most theoretical studies on data optimization aim to explain why the existing methods are effective or ineffective.

In data perception, researchers usually conducted theoretical analysis on the role of one typical data measure or leveraged the measure to understand the training process of DNNs. Doan et al.~\cite{ATA2021} conducted a theoretical analysis of catastrophic forgetting in continuous learning with neural-tangent-kernel overlap matrix. Chatterjee et al.~\cite{OTG2022} utilized the perception on gradients to explain the generalization of deep learning.

In data resampling, existing theoretical studies focus on importance sampling for deep learning. Katharopoulos and Fleuret~\cite{NAS2018} derived an estimator of the variance reduction achieved with importance sampling in deep learning. Katharopoulos and Fleuret~\cite{BISF2017} theoretically revealed that the loss value can be used as an alternative importance metric, and propose an efficient way to perform importance sampling for a deep model. Wang et al.~\cite{LIB2020} proposed an unweighted data sub-sampling method, and proved that the subset-model acquired through the method outperforms the full-set-model.

In data augmentation, more and more theoretical studies are performed. Dao et al.~\cite{AKT2019} established a theoretical framework for understanding data augmentation. According to their framework, data augmentation is approximated by two components, namely, first-order feature averaging and second-order variance regularization. Zhao et al.~\cite{MAD2020} defined an effective regularization term for adversarial data augmentation and theoretically derived it from the information bottleneck principle. Wu and He~\cite{AUFF2022} investigated the theoretical issues for adversarial perturbations for multi-source domain adaptation. Gilmer et al.~\cite{AEAA2019} also attempted to explain the adversarial samples.

In data perturbation, most theoretical studies focus on the adversarial perturbation. Yi et al.~\cite{IOG2021} investigated the models trained by adversarial training on OOD data and justified that the input perturbation robust model in pre-training provides an initialization that generalizes well on downstream OOD data. Peck et al.~\cite{LBO2017} formally characterized adversarial perturbations by deriving lower bounds on the magnitudes of perturbations required to change the classification of neural networks. Some studies delved into the theoretical justification for label and logit perturbation. Xu et al.~\cite{TUL2017} analyzed the convergence SGD with label smoothing regularization and revealed that an appropriate LSR can help to speed up the convergence of SGD. Li et al.~\cite{CLP2023} theoretically analyzed the usefulness of logit adjustment in dealing with class imbalanced issues.

In data weighting, Byrd and Lipton~\cite{WIT2019} investigated the role of importance in deep learning. Fang et al.~\cite{RIW2020} discussed the limitations of importance weighting and found that it suffers from a circular dependency. Meng et al.~\cite{ATUO2017} analyzed the capability of the self-paced learning and provided an insightful interpretation of the effectiveness of several classical SPL variations. Weinshall et al.~\cite{CLB2018} proved that the rate of convergence of an ideal curriculum learning method is monotonically increasing with the learning difficulty of the training samples. 

In data pruning, theoretical studies are relatively limited. Zhu et al.~\cite{RDD2023} revealed that distilled data lead to networks that are not calibratable. The reason lies in two folds, including a more concentrated distribution of the maximum logits and the loss of information that is semantically meaningful but unrelated to classification tasks. Dong et al.~\cite{PFF2022} emerged dataset distillation into the privacy community and theoretically revealed the connection between dataset distillation and differential privacy.

There are also studies which aim to reveal the intrinsic connections between two different technical paths. For instance, regularization is a widely used technique in deep learning~\cite{AOAS2022}, and several typical data optimization techniques are revealed to be a regularization method~\cite{ISDA2019,RKD2020,NNR2021}. Therefore, intrinsic connections among these techniques can be established, which enlightens a better understanding of the involved technical paths and can envision novel inspirations or methods. 

\begin{figure}[t] 
    \centering \includegraphics[width=1\linewidth]{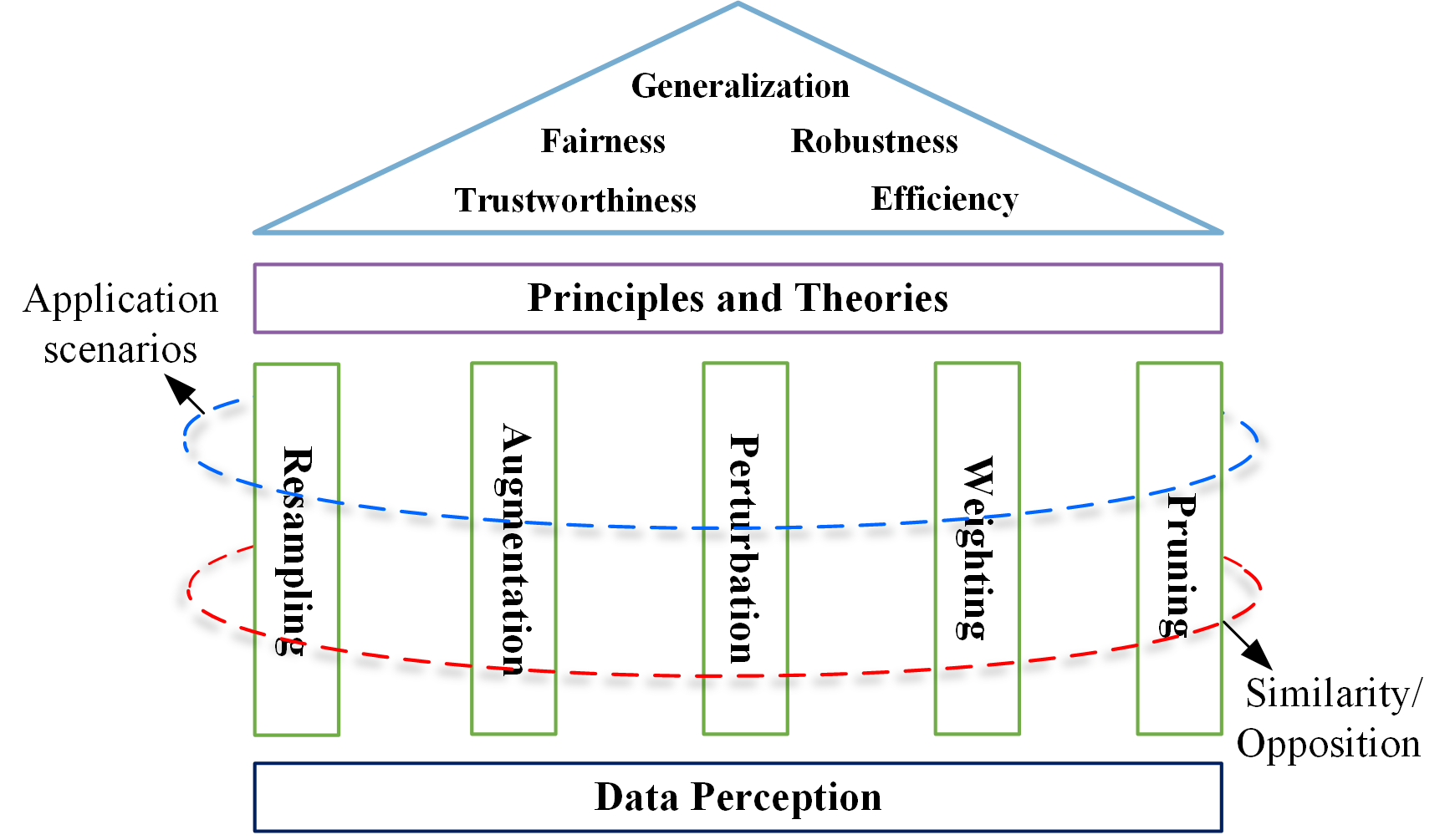}    \vspace{-0.16in}
    \caption{High-level connections for existing data optimization studies.}
    \label{fig12}
     \vspace{-0.2in}
\end{figure}

\begin{table*}[]\label{nll}
\caption{Some data optimization methods in noisy-label learning.}\vspace{-0.1in}
\centering
\begin{tabular}{|c|p{2.5cm}|p{2.8cm}|p{2.5cm}|p{2.7cm}|p{2.5cm}|}
\hline
Datasets & Resampling & Augmentation & Perturbation & Weighting & Pruning \\ \hline
CIFAR10 &~\cite{TUD2021}, ~\cite{DVUR2020},~\cite{LIB2020}&~\cite{ASFL2021},~\cite{PHS2021},~\cite{AAW2022},~\cite{DLW2020} &~\cite{RTIA2016}, ~\cite{TDNN2015}&~\cite{SLF2010},~\cite{CLA2023},~\cite{LFS2021},~\cite{DMF2020}&~\cite{DUD2019}, ~\cite{CFR2020},~\cite{PTO2022}\\ \hline
CIFAR100 &~\cite{TUD2021}, ~\cite{DVUR2020},~\cite{LIB2020} &~\cite{ASFL2021},~\cite{PHS2021},~\cite{AAW2022},~\cite{DLW2020} &~\cite{RTIA2016}, ~\cite{TDNN2015}&~\cite{SLF2010},~\cite{LFS2021},~\cite{DMF2020}&~\cite{CFR2020},~\cite{PTO2022}\\ \hline
Clothing1M & ~\cite{LIB2020}&~\cite{ASFL2021},~\cite{DLW2020},~\cite{PHS2021} &~\cite{RTIA2016}, ~\cite{TDNN2015}&~\cite{SLF2010},~\cite{DMF2020}&~\cite{PTO2022}\\ \hline
SVHN & ~\cite{BDR2020}&~\cite{RSC2021}&~\cite{CAW2023}, ~\cite{RTIA2016}, ~\cite{TDNN2015}&~\cite{SLF2010}&~\cite{CLT2022}\\ \hline
WebVision &~\cite{CSF2023}  &~\cite{DLW2020},~\cite{PHS2021}&~\cite{LWN2022} &~\cite{CLA2023},~\cite{LFS2021},~\cite{DMF2020} &~\cite{CFR2020} \\\hline
\end{tabular}
\end{table*}

\begin{table*}[]\label{ibl}
\caption{Some data optimization methods in imbalanced learning.}\vspace{-0.1in}
\centering
\begin{tabular}{|c|p{2.6cm}|p{2.6cm}|p{2.6cm}|p{2.6cm}|p{2.6cm}|}
\hline
Datasets & Resampling & Augmentation & Perturbation & Weighting & Dataset pruning \\ \hline
CIFAR10-LT &~\cite{RCR2020} &~\cite{RCR2020}, ~\cite{GAL2023}   &~\cite{LLV2021}, ~\cite{MMS2021}&~\cite{FLD2017}, ~\cite{CLB2019}& ~\cite{SMD2022}, ~\cite{ASOC2022}\\ \hline
CIFAR100-LT &~\cite{LID2019} &~\cite{IRDA2022}, ~\cite{GAL2023}&~\cite{LLV2021}, ~\cite{MMS2021}&~\cite{FLD2017}, ~\cite{CLB2019}& ~\cite{SMD2022}, ~\cite{ASOC2022}\\ \hline
iNaturallist &~\cite{LID2019} &~\cite{IRDA2022}, ~\cite{GAL2023}&~\cite{LLV2021}, ~\cite{MMS2021}&~\cite{FLD2017}, ~\cite{CLB2019}&~\cite{SMD2022}\\ \hline
ImageNet-LT &~\cite{LID2019} &~\cite{IRDA2022}, ~\cite{GAL2023}&~\cite{LLV2021}, ~\cite{MMS2021}&~\cite{FLD2017}, ~\cite{CLB2019}& ~\cite{SMD2022}, ~\cite{IDM2023}\\ \hline
\end{tabular}\vspace{-0.1in}
\end{table*}

\section{Connections among different techniques}
The connections among different data optimizations techniques can be described by Fig.~\ref{fig12}. Four aspects, namely, perception, application scenarios, similarity/opposition, and theories, connect different methods within a technical path or across different paths. 

\subsection{Connections via data perception}
Data perception is the first~(explicit or implicit) step in the data optimization pipeline. Methods along different technical paths introduced in Section VI may choose the same or similar quantities in perception. Therefore, quantities for data perception connect different methods. For example, many data optimization methods are on the basis of training loss in resampling~\cite{PAA2020}, augmentation~\cite{SDA2021}, perturbation~\cite{LP2022}, weighting~\cite{ATUO2017}, and subset selection~\cite{MAU2023}. Gradient is widely used in resampling~\cite{NAS2018}, augmentation~\cite{MAG2023}, perturbation~\cite{CAW2023}, weighting~\cite{GHS2019}, and dataset distillation~\cite{DCWG2021}. Other quantities including margin and uncertainty are also used in different optimization techniques.

The utilization of the same or similar perception quantities demonstrates that these methods have the same or similar heuristic observations or theoretical inspirations.

\subsection{Connections via application scenarios}
Most data optimization methods can be leveraged for the application scenarios discussed in Section IV-B. 

One of the most focused scenarios of data optimization methods is  noisy-label learning. 
Many classical methods are from resampling~\cite{RNC2020}, augmentation, weighting, or perturbation. These are also dataset distillation studies for noisy-label datasets~\cite{GDD2021}. Table I shows some representative data optimization methods for noisy-label learning on five benchmark datasets CIFAR10~\cite{LML2009}, CIFAR100~\cite{LML2009}, Clothing1M~\cite{LFM2015}, SVHN~\cite{RDI2011}, and WebVision~\cite{WDV2017}.

Imbalanced learning is also among the most focused scenarios. Nearly all the listed data optimization technical paths have been used in imbalanced learning. Table II shows some representative data optimization methods for imbalanced learning on four benchmark datasets CIFAR10-LT~\cite{CLB2019}, CIFAR100-LT~\cite{CLB2019}, iNaturalist~\cite{TIS2018}, and ImageNet-LT~\cite{LLR2019}. There are some studies employing more than one type of data optimization techniques such as ReMix~\cite{RCR2020}, which combines resampling and augmentation, in Table II.

Robust learning for adversarial attacks is another typical scenario. Karimireddy and Jaggi~\cite{BLH2022} employed resampling to design robust model in distributed learning. Data weighting~\cite{GIA2021} and dataset distillation~\cite{CWA2022} are also used in robust learning.

\subsection{Connections via similarity/opposition}
The similar and opposite relationships existing among the five technical paths are introduced in Section VI.

Data resampling and weighting are closely related techniques, as their key steps are nearly the same. Therefore, in many studies on noisy-label learning and imbalanced learning, these two techniques are often considered as a single strategy.

\begin{figure}[ht] 
    \centering \includegraphics[width=0.9\linewidth]{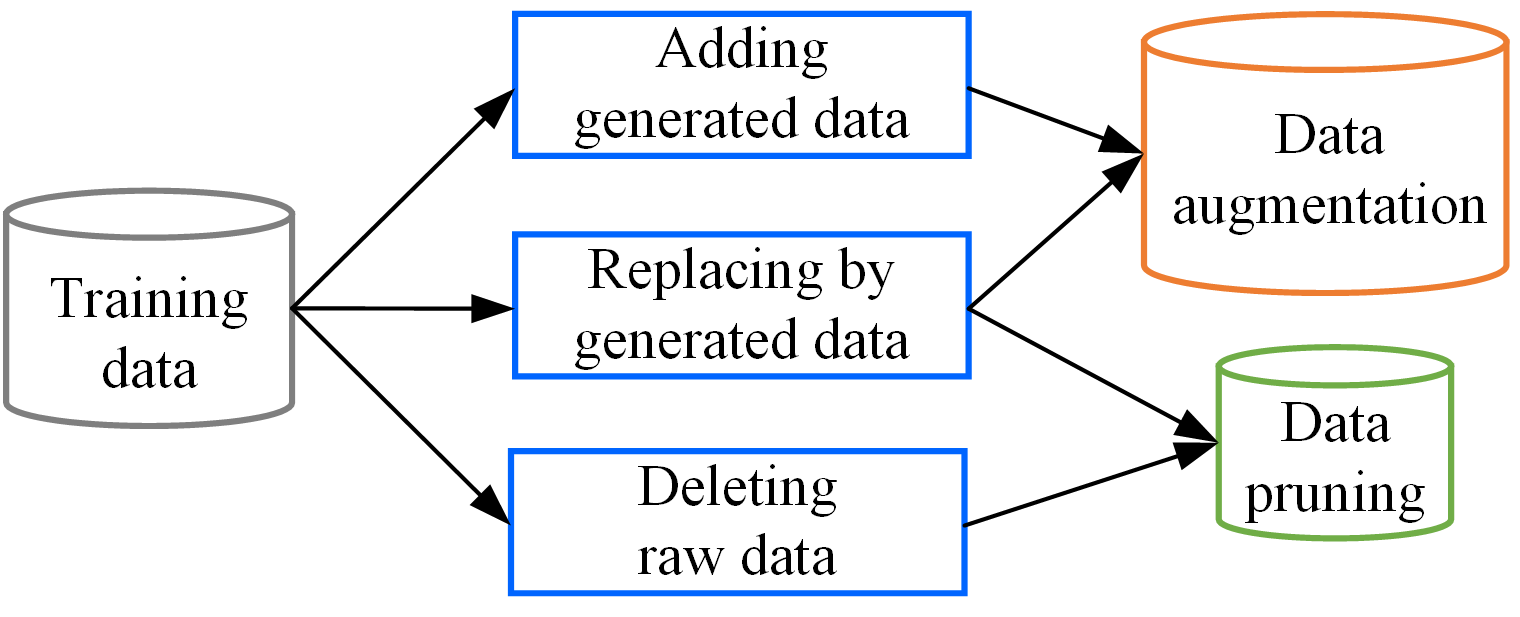}    \vspace{-0.08in}
    \caption{Connection between augmentation and pruning.}
    \label{fig13}
     \vspace{-0.02in}
\end{figure}

Although data pruning and augmentation are opposite to each other, they have consistent ultimate goals in learning tasks. They are overlapped in terms of employed methodologies as shown in Fig.~\ref{fig13}. It is believable that more intrinsic connections can be explored for them.

In the data resampling, weighting, and perturbation, the assignment manners for the sampling rate, weighting score, and perturbation variable are quite similar. In addition to the classical importance score, both meta learning~\cite{MBE2020} and adversarial strategy~\cite{BDR2020} have also been used in data resampling. Regularization-based manner is used in nearly all the data optimization paths except resampling. Due to space constraints, methods with different assignment manners are not summarized in a table as those in Section VIII-B.

There are other opposite relationships, such as undersampling vs. oversampling, easy-first weighting vs. hard-first weighting, positive perturbation vs. negative perturbation, and explicit augmentation vs. implicit augmentation. Both methodologies in these opposite relationships have been demonstrated to be effective, aligning with the proverb ``All roads lead to Rome".

\subsection{Connections via theory}
There are some common theoretical issues, analyses, and conclusions heavily influencing most data optimization techniques. They are the natural connections among different techniques. Several examples are listed as follows:
\begin{itemize}
    \item Theoretical issues in data perception. A solid theoretical basis for data perception in data optimization is lacking, even though most data optimization methods implicitly or explicitly rely on the perception for the training data. For instance, many methods from resampling, weighting, and perturbation are based on dividing samples into easy and hard. Nevertheless, there is not yet a widely accepted learning difficulty measure with a rigorous theoretical basis in the literature. More than ten types of learning difficulty measures are utilized to distinguish easy from hard samples in previous literature. A theoretical formulation for data perception is of great importance.
    \item Probabilistic density (ratio) estimation. Many data optimization methods, especially data resampling and weighting, heavily rely on the probabilistic density (ratio) estimation. The most representative method is the importance sampling. In learning difficulty-based weighting, the probabilistic density ratio, in terms of learning difficulty, is revealed to determine the priority mode~\cite{WSS2023}, namely, easy/medium/hard-first.
    \item Regularization-based explanation. Many data optimization methods are considered as a type of regularization, including data augmentation and perturbation. In these methods, data optimization performs implicit model regularization other than explicit regularization that directly works on model parameters. Regularization is not always beneficial as over-regularization may occur. Li et al.~\cite{AND2019} pointed out that large amount of augmented noisy data could lead to over-regularization and proposed an adaptive augmentation method. Adversarial training may result in robust overfitting~\cite{URO2022}.
    \item Generalization bound for data optimization. Many studies choose to deduce a mathematical bound for the generalization risk in terms of the empirical risk and variables related to the data optimization. This manner can theoretically explain the utility of the involved data optimization. Xiao et al.~\cite{SAAG2022} derived stability-based generalization bounds for stochastic gradient descent (SGD) on the loss with adversarial perturbations. Xu et al.~\cite{UTR2021} established a new generalization bound that reflects how importance weighting leads to the interplay between the empirical risk and the deviation between the source and target distributions. 
\end{itemize}

The progress in each of the above theoretical aspects will promote the advancement of many data optimization methods in different technical paths. Hopefully, this survey will promote the mutual understanding of the referred technical paths.

\section{Future directions}
This section summarizes some research directions deserving further exploring.

\subsection{Principles of data optimization}

Up till now, there has been no consensus theoretical framework that is suitable for all or most technical paths. There are some studies aiming to establish the connection between two different technical paths, such as resampling vs. weighting~\cite{WRO2021}. Many open problems or controversies remain unsolved. For example, there is no ideal answer for which resampling strategy should be employed first: oversampling or undersampling? Megahed et al.~\cite{TCI2021} suggested that undersampling should be used firstly, whereas Xie et al.~\cite{GDB2022} demonstrated that oversampling is effective. Likewise, although Zhou et al.~\cite{WSS2023} provided an initial answer for the choice of easy-first and hard-first weighting strategies, a solid theoretical framework is still lacking in their study. 

Moreover, even for a single data optimization method, multiple explanations from different views may exist. The explanation for label smoothing is a typical example. At least four studies provide empirical or theoretical explanations for it~\cite{TUL2017,WDL2019,AIO2020,GER2020}. Regarding the effectiveness of adversarial samples, some researchers have pointed out that adversarial samples are useful features~\cite{AEA2019}, while some other researchers investigated it in terms of gradient regularization~\cite{ASOR2022}. 

Consequently, the construction of the data optimization principles is of great importance, as it can promote the establishing of a unified and solid theoretical framework which can be used to analyze and understand of each data optimization technical path. There have been studies on the first principle for the design of DNNs~\cite{RAW2022}. To explore the principles for data optimization, a unified mathematical formalization tool is required and large-scale empirical studies~(e.g., \cite{CEA2021}, \cite{DLM2020}) will also be helpful.

\subsection{Interpretable data optimization}

Interpretable data optimization refers to the explanation for the involved data optimization techniques in terms of how and which aspects they affect the training process of DNNs. Although interpretable deep learning receives much attention in recent years~\cite{IDL2022}, it focuses on DNN models other than the training processing in which data optimization techniques are involved. Interpretable data optimization is an under-explored research topic and there are limited studies on this topic~\cite{CPG2023}. The well explanation of how and which aspects of a data optimization method affects a specific training process is significant beneficial for the design or selecting of more effective optimization methods.

The aforementioned theoretical studies on data optimization provide partial explanations for the corresponding data optimization method. However, the partial explanations are concentrated in common aspects across different learning tasks. How and which aspects for the involved method on a specific learning task remain unexplored.

The interpretable deep learning area has raised many effective methodologies. Recently, researchers have attempted to introduce interpretable methodologies  to explore the data optimization methods. Zelaya and Vladimiro~\cite{TET2019} explored metrics to quantify the effect of some data-processing steps such as undersampling and data augmentation on the model performance. Hopefully, more and more studies on explainable data optimization appear in the future.

\subsection{Human-in-the-loop data optimization}
Recently, human-in-the-loop~(HITL) deep learning receives increasing attention in the AI community~\cite{HML2023}. With out human's participants, high-quality training samples are not intractable to obtain. Naturally, HITL data optimization can also be beneficial for deep learning. Collins et al.~\cite{HMI2023} investigated HITL mixup and indicated that collating humans’ perceptions on augmented samples could impact model performance. Wallace et al.~\cite{TMI2019} proposed HITL adversarial generation, where human authors are guided to break models. Agarwal et al.~\cite{EED2022} proposed Variance of Gradients (VoG) to measure samples' learning difficulty and ranked samples by VoG. Then, a tractable subset of the most difficult samples is selected for HITL auditing. Overall, research on HITL data optimization is in the early stage.

\subsection{Data optimization for new challenges}
New challenges are constantly emerging in deep learning applications. We take the following three recent challenges as examples to illustrate the future direction of data optimization:
\begin{itemize}
   \item Open-world learning. This learning scenario confronts the challenge of out-of-distribution~(OOD) samples. Wu et al.~\cite{NAU2021} investigated noisy-label learning under the open-world setting, in which both OOD and noisy samples exist. Some other studies investigate cases when ODD meets imbalanced learning~\cite{ICL2021} and adversarial robustness~\cite{GBN2022}. 
   \item Large-model training. Large models especially the large language models~\cite{ASOL2023} have achieved great success in recent years. Data optimization can also take effect in the training of large models. Wei et al.~\cite{IA22023} investigated the condensation of prompts and promising results are obtained. Contrarily, Jiang et al.~\cite{CLM2023} leveraged prompt augmentation to calibrate large language models. Many issues investigated in conventional deep learning tasks may also exist for large-model training, e.g., prompt valuation.
    \item Multi-modal learning. With the development of data sensing and collection technology, multi-modal data are avaliable in more and more real tasks~\cite{MML2019}. Consequently, many learning tasks are actually multi-modal learning. As each sample consists of raw data/features from different modalities, the data perception for multi-modal samples should be different from that for conventional single-modal samples. The data optimization methods are likewise different from conventional methods~\cite{LRM2023}.
\end{itemize}

\subsection{Data optimization agent}

Given a concrete learning task, a selection dilemma occurs for the tremendous data optimization techniques. There have been studies on the automatic data optimization such as automatic data augmentation~\cite{ADA2023}. Nevertheless, existing automatic data optimization methods still focus on a particular type of technical path rather than the types across different technical paths~\cite{IAD2022,AOP2021}. A more general data optimization agent can be trained by iteratively training on a large number of deep learning tasks via reinforcement learning.

Fig.~14 shows a possible mean to construct a data optimization agent. New learning tasks are compiled based on existing classical tasks via operations such as noise adding, and class proportion re-distributing. The candidate of data optimization operators are from arbitrary optimization techniques or a single technique introduced in Section VI. A powerful data optimization agent can then be trained via reinforcement learning based on compiled learning tasks and their rewards. 

\begin{figure}[t] 
    \centering \includegraphics[width=0.98\linewidth]{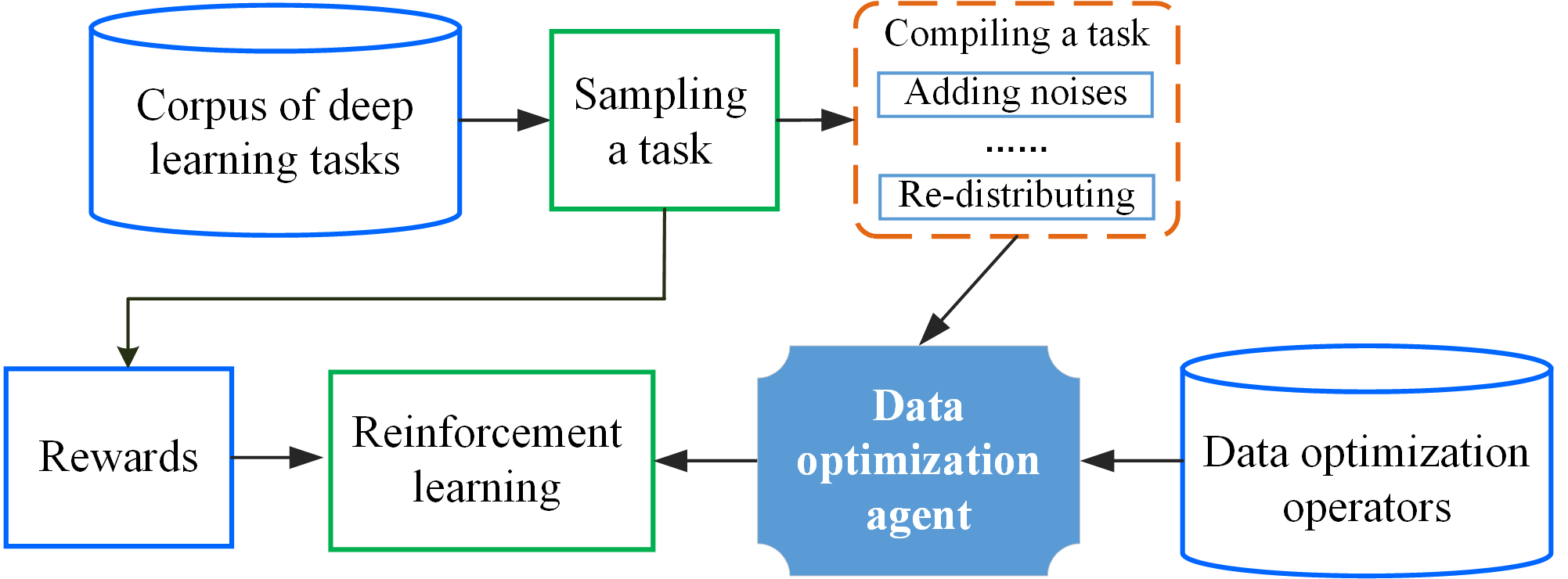}    \vspace{-0.08in}
    \caption{The training of a data optimization agent.}
    \label{fig3}
     \vspace{-0.0in}
\end{figure}

\section{Conclusions}
This paper aims to summarize a wide range of learning methods within an independent deep learning realm, namely, data optimization. A taxonomy for data optimization, as well as fine-granularity sub-taxonomies, is established for existing studies on data optimization. Connections among different methods are discussed, and potential future directions are presented. It is noteworthy that many classical methods, such as dropout, are essentially data optimization methods. In our future work, we will explore a more fundamental and unified viewpoint on data optimization, and develop a more comprehensive taxonomy to incorporate more classical methods. We hope that this study can inspire more researchers to gain insight into data-centric AI.

\bibliography{ReferenceFormat}
\bibliographystyle{IEEEtran}

\vfill

\end{document}